\documentclass[sigconf]{acmart}
\usepackage{multirow}
\usepackage{balance}
\usepackage{bbding}
\AtBeginDocument{%
  }

\begin{document}

\title{Degeneration-Tuning: Using Scrambled Grid shield Unwanted Concepts from Stable Diffusion}

\author{Zixuan Ni}
\authornote{Work done when interning at Huawei Cloud.}
\email{zixuan2i@zju.edu.cn}
\orcid{0000-0001-5196-983X}
\affiliation{%
  \institution{Zhejiang University}\country{China}
}

\author{Longhui Wei}
\email{weilonghui@huawei.com}
\orcid{0000-0001-6916-3009}
\affiliation{%
  \institution{Huawei inc.}\country{China}
}

\author{Jiacheng Li}
\email{lijiacheng@zju.edu.cn}
\orcid{0000-0003-2964-6107}
\affiliation{%
  \institution{Zhejiang University}\country{China}
}

\author{Siliang Tang}
\email{ siliang@zju.edu.cn}
\orcid{0000-0002-7356-9711}
\affiliation{%
  \institution{Zhejiang University}\country{China}
}

\author{Yueting Zhuang}
\authornotemark[2]
\email{yzhuang@zju.edu.cn}
\orcid{0000-0001-9017-2508}
\affiliation{%
  \institution{Zhejiang University}\country{China}
}

\author{Qi Tian}
\email{tian.qi1@huawei.com}
\orcid{0000-0002-7252-5047}
\affiliation{%
  \institution{Huawei inc.}\country{China}
}
\authornote{Corresponding author.}
\renewcommand{\shortauthors}{Zixuan Ni et al.}
\newcommand{\nzx}[1]{\textcolor{red}{[#1]}}
\begin{abstract}
Owing to the unrestricted nature of the content in the training data, large text-to-image diffusion models, such as Stable Diffusion (SD), are capable of generating images with potentially copyrighted or dangerous content based on corresponding textual concepts information. This includes specific intellectual property (IP), human faces, and various artistic styles. However, Negative Prompt, a widely used method for content removal, frequently fails to conceal this content due to inherent limitations in its inference logic. In this work, we propose a novel strategy named \textbf{Degeneration-Tuning (DT)} to shield contents of unwanted concepts from SD weights. By utilizing Scrambled Grid to reconstruct the correlation between undesired concepts and their corresponding image domain, we guide SD to generate meaningless content when such textual concepts are provided as input. As this adaptation occurs at the level of the model's weights, the SD, after DT, can be grafted onto other conditional diffusion frameworks like ControlNet to shield unwanted concepts. 
In addition to qualitatively showcasing the effectiveness of our DT method in protecting various types of concepts, a quantitative comparison of the SD before and after DT indicates that the DT method does not significantly impact the generative quality of other contents. The FID and IS scores of the model on COCO-30K exhibit only minor changes after DT, shifting from 12.61 and 39.20 to 13.04 and 38.25, respectively, which clearly outperforms the previous methods.
\end{abstract}



\maketitle

\begin{figure*}[!t]
  \includegraphics[width=\textwidth,height=0.51\textwidth]{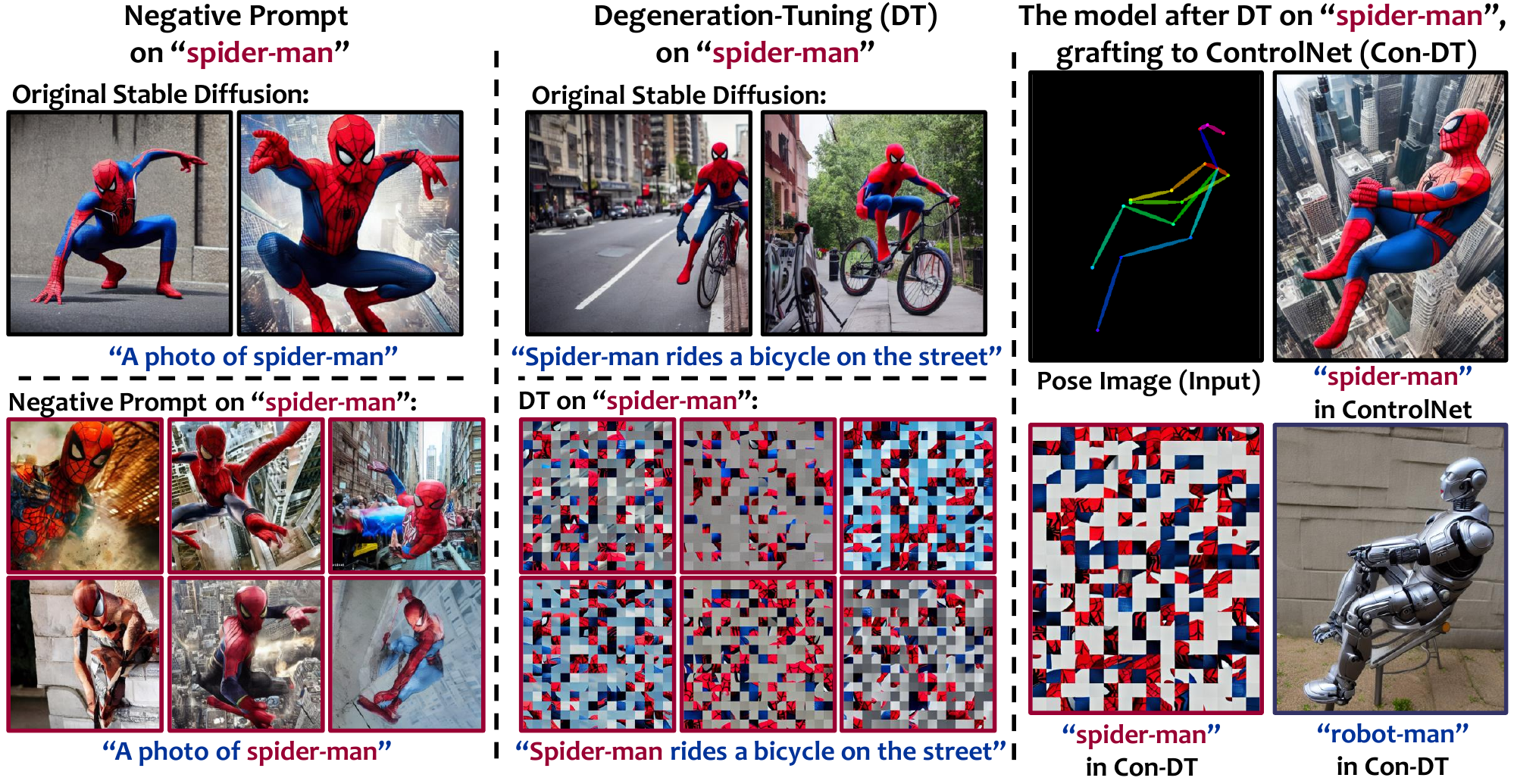}
  \vspace{-2em}
  \caption{The left sub-figure illustrates the limitations of the popular Negative Prompt method in concealing specific concepts.  The middle sub-figure shows the effectiveness of our Degeneration-Tuning (DT) method in shielding specific concepts. The right sub-figure demonstrates that the model, after DT on unwanted concepts, remains effective when grafted it into other conditional diffusion model like ControlNet\cite{zhang2023adding} (Con-DT).
  }
  \label{first_image}
  \vspace{-1em}
\end{figure*}
\section{Introduction}

Large-scale text-to-image diffusion models \cite{dalle2,rombach2022high,nichol2021glide,saharia2022photorealistic,liu2023cones,liu2023cones2} such as Stable Diffusion (SD) \cite{rombach2022high}, have garnered significant attention from both industry and media. However, they have also given rise to various social issues, such as potential infringement of intellectual property (IP) \cite{dalle2}, inappropriate use of human faces, and misuse of various artistic styles \cite{qisu}. Furthermore, these images could be used to spread misleading information or rumors about celebrities and politicians, damaging their reputations and disrupting social harmony \cite{fakenews}. The primary reason behind these issues is that the data used to train these diffusion model is unrestricted and often contains inherent human biases. 

Recently, three strategies have been produced to prevent the SD model from generating undesirable content. The first involves limiting the training contents \cite{opensd20}. The second aims to disturb inference (generation) process of the SD model \cite{rombach2022high}, while the third strategy employs a Safety Filter \cite{rando2022red}. The most effective method within the first strategy is to filter out undesired contents from the training data \cite{rombach2021highresolution}. However, 
removing this sensitive/unwanted content from large scale training datasets such as LAION-5B \cite{LAION} and retraining a SD model from them always consumes over 150,000 GPU-hours \cite{sd14}. Moreover, sensitive or infringing content is not static and changes over time, making it impractical to retrain the model merely for a few emerging concepts. Although negative prompt method \cite{NP}, a popular methods within the second strategy, can remove unwanted content from the generation process, it's not universally effective. As shown in Figure \ref{first_image}, when the input prompts resemble the negative prompts, the generated images often retain negative prompt information. The Safety Filter strategy filters out the generated images which activate pre-trained special content classifiers. However, as mentioned in \cite{rando2022red}, these classifiers often become obfuscated and exhibit poor precision when test samples fall outside the training data domain. Additionally, continually adding safety filters not only complicates the entire generation framework but also significantly slows down the generation feedback speed. Furthermore, these filters can be easily circumvented \cite{remove_filter}. Another potential risk is that the latter two methods only influence model's inference and output. If the parameters of SD are attacked or leaked, these strategies will be bypassed.

\begin{figure*}[!h]
  \includegraphics[width=\textwidth,height=0.22\textwidth]{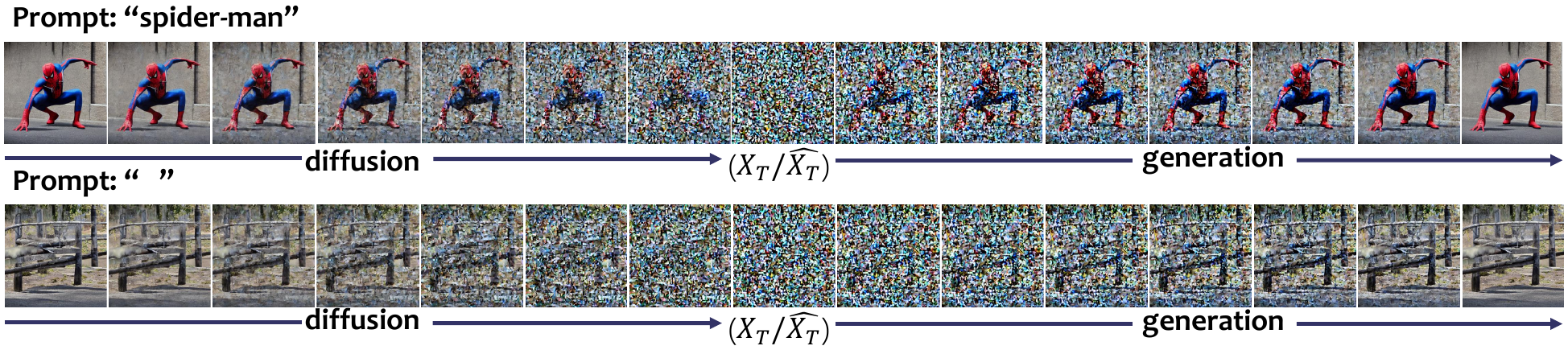}
  \vspace{-2em}
  \caption{This figure presents the diffusion and generation process of the stable diffusion under fixed random seeds, with and without conditional information ("spider-man" or " "). Each sub-figure represents the processing result after equal time step $\tau$. It's evident that low-frequency signals or longer wavelength features in the diffusion process are the last to disappear, while they are the first to appear in the generation process. And the conditional input (prompt) influences the low-frequency information in the generation process.}
  \label{diffusion_and_generation}
  \vspace{-0.5em}
\end{figure*}

In this paper, we analyze the generative mechanism of the diffusion model and discover that the primary factor influencing the model's semantic content generation is the distance between the initial sampled Gaussian noise and the final diffusion distribution within the training data domain. Moreover, the conditional information in diffusion model is responsible for predicting or correcting the distance between the distribution of the current denoised samples and the current diffusion distribution within the specific content's training data domain. Observing the diffusion and generation processes of the samples with or without prompts (Figure \ref{diffusion_and_generation}), we note that the conditional information initially steers the low-frequency features of the image content during the generation process. These features are also the last to disappear during the training (diffusion) process. Consequently, we deduce that the primary components of this distribution distance are the low-frequency signals, and SD model has learned the low-frequency image contents associated with various linguistic concepts.
Inspired by these insights, we propose a novel method called \textbf{Degeneration Tuning(DT)} to shield unwanted concepts from SD model. By employing Scrambled Grid operation to disrupt the low-frequency visual content of the conditional concepts, we construct a degraded dataset. After re-tuning the SD model on these dataset, we reconstruct the model's predictions for the visual contents associated with unwanted concepts. Importantly, due to re-tuning, our DT method shields specific concepts at the level of model's parameters, it remains effective even if the model parameters are leaked. Figure \ref{first_image} briefly showcases the effectiveness of DT method. In Section \ref{experiments}, we qualitatively demonstrate that DT method can accurately protect various types of concepts in different continual contexts. The diffusion module (U-net network \cite{ronneberger2015u,huang2021multi,huang2022singgan}) in stable diffusion after DT can be grafted into other condition-controlled diffusion models, such as ControlNet \cite{zhang2023adding}. Quantitative analysis in Section \ref{evaluation} further proves that DT method, while shielding specific concepts, does not significantly impact the model's generation ability to generate general content. After DT, the model's FID and IS scores on the COCO-30K dataset are 13.04 and 38.25, clearly surpasses previous Erase \cite{gandikota2023erasing} and SLD \cite{schramowski2023safe} methods. 

The contribution of this work are summarized as: 
\begin{itemize}
    \item We analyzed the generative mechanism of diffusion models and discovered that the primary factor influencing the model's semantic content generation is the distance between the initial sampled Gaussian noise and the final diffusion distribution within the training data domain.
    \item We proposed a simple yet effective method called \textbf{Degeneration Tuning} to protect specific content in stable diffusion. Quantitative and qualitative results presented in Section \ref{experiments} and \ref{evaluation} illustrate the effectiveness of our DT method in shielding specific concepts without harming other content.
    \item We further valid the feasibility and challenges of continual DT for its future online applications.
\end{itemize}

\section{Related Work}
\subsection{Conditional Diffusion Probabilistic Models}
With the emergence of Denoising Diffusion Probabilistic Model (DDPM) \cite{ho2020denoising}, Denoising Diffusion Implicit Model (DDIM) \cite{song2020denoising} and score-based diffusion \cite{song2021maximum}, the quality of image generated by diffusion probabilistic model (DM) \cite{sohl2015deep} has been improved and the inference time of the model also has been shortened. What's more, classifier-guidance \cite{dhariwal2021diffusion} and classifier-free \cite{ho2022classifier,dalle2,rombach2022high,kawar2022imagic,sun2023sddm} strategies make DMs can generate specific contents based on classifier information rather than initial Gaussian noise. In order to reduce the computation power required for training a diffusion model, based on the idea of latent features \cite{karras2019style,yan2021videogpt,yu2021vector,esser2021taming,wei2022mvp,li2022fine,ni2023continual}, the approach Latent Diffusion Model (LDM) \cite{rombach2022high} was proposed and further extended to Stable Diffusion \cite{sd14}. 
Beside of training more effective conditional diffusion models, there are also some works fine-tuning stable diffusion model using specific data \cite{gal2022image,hu2021lora,kim2022diffusionclip,hertz2022prompt}. DreamBooth \cite{ruiz2022dreambooth} try to use self-definition text embedding [V] to teach stable diffusion model generate private and fix images. By freezing the parameters of the original stable diffusion model and embedding a new module which is fine-tuned in specific conditional datasets, ControlNet \cite{zhang2023adding} and T2I-Adapter \cite{mou2023t2i} control the output of the stable diffusion using their wanted conditional information. 
Although these methods can guide stable diffusion to generate specific contents based on new added conditional information, they do not consider how to shield or erase some contents for which the conditional information is already known or exists. 

\subsection{Limitation in Negative Prompt Method}
As the most widely applied method for removing unwanted contents, Negative Prompt (NP) \cite{sd14} has been used in various diffusion models. However, this method faces an unavoidable issue. When the input prompts resemble the negative prompts, the generated images often retain negative prompts information. The reason is that the inference process of negative prompt method is:
\begin{equation}
\epsilon_{\theta}(x_t,t,c,c_{NP}) = \epsilon_{\theta}(x_t,t,c) + \lambda * (\epsilon_{\theta}(x_t,t,c)-\epsilon_{\theta}(x_t,t,c_{NP})).
\end{equation}
where the $c$ is positive prompts, $c_{NP}$ is negative prompts and $\lambda$ is hyper-parameter. When the negative prompts equals positive prompts, there must exist item $\epsilon_{\theta}(x_t,t,c)$ regardless of what $\lambda$ is. The concrete examples can be seen in Figure \ref{first_image}.

\subsection{Continual Learning}
Continual learning \cite{mccloskey1989catastrophic} is a learning paradigm that training the model based on current data and the past data are unavailable. The core challenge of continual learning is to enable the model to continuously learn new knowledge while preserving previously learned knowledge, without experiencing catastrophic forgetting \cite{rebuffi2017icarl,kirkpatrick2017overcoming}, which can lead to a decline in performance on previously learned tasks. This challenge is difficult because the past data cannot be utilized \cite{rebuffi2017icarl,goodfellow2013empirical,ni21revisit,ramasesh2020anatomy}. It also means that when the model updates its parameters based on existing data, it will not be limited by the past data domain. This unavailable data plays an important role in catastrophic forgetting \cite{kirkpatrick2017overcoming}. However, in generative tasks, this unavailable data can be generated from the generative model itself \cite{carlini2023extracting}, although the quality of this is not as good as the original data. This provides the possibility for continual learning without catastrophic forgetting in generative tasks.

\begin{figure*}[!t]
  \includegraphics[width=\textwidth,height=0.33\textwidth]{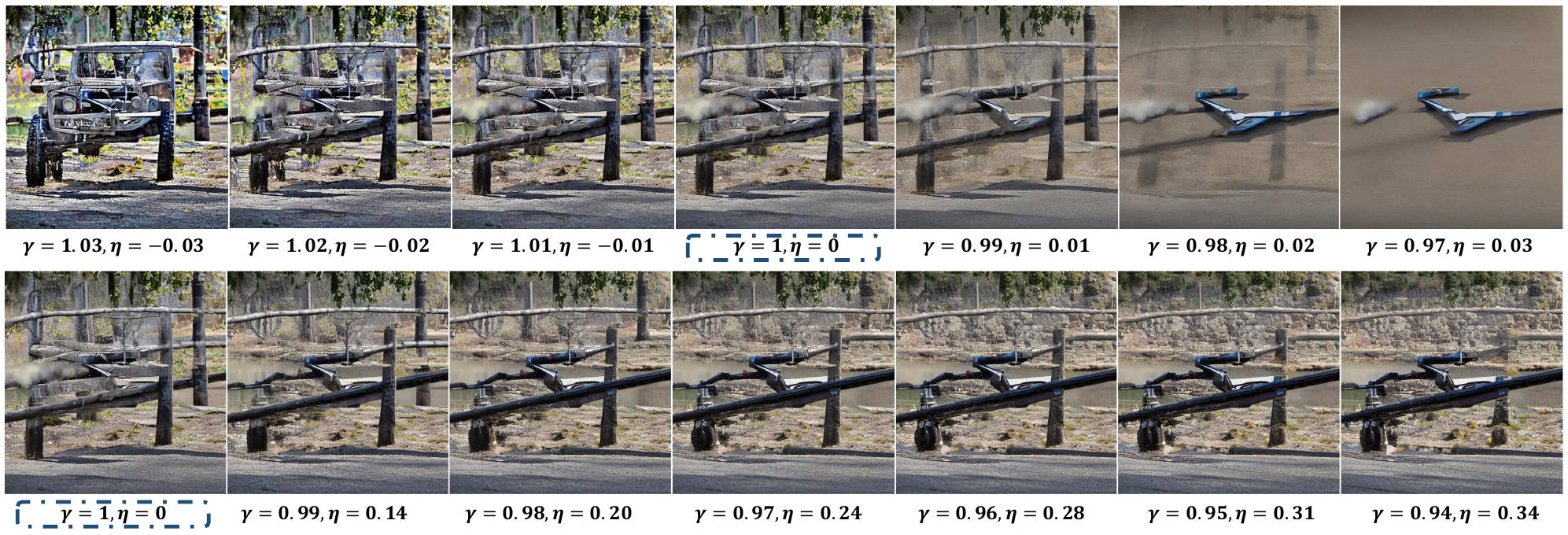}
  \vspace{-2em}
  \caption{The results are generated by the stable diffusion with a fixed random seed and None conditioanl information(" "). By resetting the initial noise $\hat{x}_T = \gamma \epsilon + \eta \epsilon_0$ and slightly adjusting the coefficients $\gamma$ and $\eta$ based on the value of the $\hat{x}_T$ (the first row) or the distribution of the $\hat{x}_T$ (the second row), we can conclude that small variations in the distribution of initial noise significantly affect the semantic information of the images generated by the diffusion model.}
  \label{compare_frequence}
  \vspace{-1em}
\end{figure*}

\section{Methods}
\label{methods}
\subsection{Preliminaries}
\label{Preliminaries}
\textbf{Diffusion models(DM)}, as the most promising generation strategies currently, aim to learn a data distribution $p(x)$ by progressively denoising a random variable sampled from a Gaussian distribution $\mathcal{N}\sim(0,\mathcal{I})$. In the diffusion process, by continually adding Gaussian noise $\epsilon$ to the image $x_0$ sampled from $p(x)$, the diffusion model learns the relationship between the data distribution $p(x)$ and Gaussian distribution $\mathcal{N}\sim(0,\mathcal{I})$. The formula in diffusion process is:
\begin{equation}
x_t = \alpha_t x_0 + \sigma_t\epsilon_t,
\end{equation}
where the time steps $t = 1,...,T$. The $\alpha_t$ is a decreasing functions based on $t$ which close to 0 when $t$ come to $T$. The $\sigma_t=\sqrt{1-\alpha_t^2}$, and $\epsilon_t$ represents the Gaussian noise for different time steps $t$, sampled from $\mathcal{N}\sim(0,\mathcal{I})$. The diffusion model, denoted as $\epsilon_{\theta}(x_t,t)$, is trained to predict noise $\epsilon_t$ based on the time step $t$ and $x_t$. The training loss $L_{DM}$ \cite{ho2020denoising} can be simplified to:
\begin{equation}
L_{DM} := \mathbb{E}_{x,t,\epsilon _t \sim \mathcal{N}(0,1)}[ ||\epsilon _t - \epsilon_{\theta}(x_t,t)||_2^2 ],
\end{equation}

To enable DM training on limited computational resources while retaining their quality and flexibility, the Latent Diffusion Model (LDM) \cite{rombach2021highresolution} employs powerful pre-trained auto-encoders $\varepsilon$, such as VAE \cite{kingma2013auto,rezende2014stochastic,esser2021taming,frans2022clipdraw}, to encode image data $x$ into the latent space $z$, where $z = \varepsilon (x)$. Additionally, by introducing cross-attention layers into the model architecture, LDM can generate image content based on input conditional information, such as text or bounding boxes. Based on this image-condition pairs, the loss of conditional ldm $L_{LDM}$ can be designed as:
\begin{equation}
L_{LDM} := \mathbb{E}_{\varepsilon (x), y, t, \epsilon _t \sim \mathcal{N}(0,1)}[ ||\epsilon _t - \epsilon_{\theta}(z_t,\tau_{\theta}(y), t)||_2^2 ],
\end{equation}
where $y$ and $\tau_{\theta}$ are the conditional inputs and its encoders.

\subsection{Motivation}
\label{Motivation}
As described in Section \ref{Preliminaries}, the diffusion process in DM is $x_{t-1} = \alpha_{t-1} x_0 + \sigma_{t-1}\epsilon_{t-1}, \epsilon_{t-1}\sim\mathcal{N}(0,\mathcal{I})$, which can also be written as: 
\begin{equation}
P(x_{t-1}|x_{0}) = \mathcal{N}(x_{t-1}, \alpha_{t-1} x_0,\sigma_{t-1}^2\mathcal{I}) 
\end{equation}
The reverse or generation process can be summarized as: 
\begin{equation}
P(x_{t-1}|x_{t}) = \mathcal{N}(x_{t-1}; \mu_{\theta}(x_t,t), \sigma_t^2\mathcal{I}) 
\end{equation}
where the $\mu_{\theta}(x_t,t)$ is diffusion model to predict $x_0$ based on $x_t$ and time step $t$. With the increasing of time step $t$, the $x_t$ is \textbf{getting closer} to $\mathcal{N}\sim(0,\mathcal{I})$ but \textbf{never equal} to $\mathcal{N}\sim(0,\mathcal{I})$. However, in practical generation process, we sample $\hat{x}_T$ from $\mathcal{N}\sim(0,\mathcal{I})$ to replace diffusion result $x_T$ as the initial input in generation process. We can formulate the difference between $\hat{x}_T$ and $x_T$ as:
\begin{equation}
x_T = \hat{x}_T + \Delta ,
\end{equation}
where the $\Delta$ is a small amount. Then, what role does $\Delta$ play in the generation process of the diffusion model? We fix the random seeds of the SD and formulate small difference of $x_T$ as :
\begin{equation}
\alpha_{T} x_0 + \sigma_{T}\epsilon = x_T = \hat{x}_T + \Delta = \epsilon + \Delta \approx \gamma \epsilon + \eta \epsilon_0, \epsilon,\epsilon_0\sim\mathcal{N}(0,\mathcal{I}),
\end{equation}
As shown in Figure \ref{compare_frequence}, when we slightly adjust the coefficients $\gamma$ and $\eta$ based on the value of the $\hat{x}_T$ (refer to the first row in Figure \ref{compare_frequence}), the generated images exhibit a noticeable change in their semantic content. However, when we adjust the coefficients $\gamma$ and $\eta$ to maintain the invariability of the distribution of $\hat{x}_T$ (refer to the the second row in Figure \ref{compare_frequence}), the semantic content of the generated images changes slightly. \textbf{The results of this is completely different from before.} Reviewing the first experiments from the distribution level, we observe that a slightly shift in the distribution of $\hat{x}_T$ from $\mathcal{N}(0,\mathcal{I})$ to $\mathcal{N}(0,0.98*\mathcal{I})$ (where $(\gamma,\eta = 1,0)$ shifts to $(\gamma,\eta = 0.98,0.02)$) causes a noticeable change in the generated content. Similarly, when the distribution of $\hat{x}_T$ shifts from $\mathcal{N}(0,\mathcal{I})$ to $\mathcal{N}(0,1.03*\mathcal{I})$ (where $(\gamma,\eta = 1,0)$ shifts to $(\gamma,\eta = 1.03,-0.03)$), the semantic content of the generated images changes again. All of this suggests that the distribution distance determines the content of the generated images.
Specifically, the contents of the images generated by a diffusion model depend on the training data domain, in which the diffusion result $x_T$ is closest to $\hat{x}_T$.


Based on this observation, we infer that the small value of $\Delta$ affects the output of the diffusion model by influencing the sampling distribution. Furthermore, what does the distribution represent in an image? We visualize the generation and diffusion process of the stable diffusion using the same Gauss Noise $\hat{x}_T$, with or without a prompt "spider-man", as illustrated in Figure \ref{diffusion_and_generation}. Each sub-image represents the processing result after an equal time step $\tau$. It's obviously that the low-frequency signals or the longer wavelength features of the results in the diffusion process are the last to disappear and are harder to be altered. In the generation process, these features are the first to appear and are still harder to be altered. Because of this, we infer that $\Delta$ implicitly maps low-frequency information of the specific data domain. And when using classifier-guidance \cite{dhariwal2021diffusion}, the generation process can be written as:
\begin{equation}
P(x_{t-1}|x_{t},y) = \mathcal{N} (x_{t-1}, \mu_{\theta}(x_t,t)+\sigma_t^2\bigtriangledown_{x_t}\log P(y|x_t),\sigma_{t}^2\mathcal{I}), 
\end{equation}
where the $\sigma_t^2\bigtriangledown_{x_t}\log P(y|x_t)$ can be considered as a correction to the specific $\Delta$ which represent a specific data domain. The classifier-free strategy \cite{ho2022classifier,dalle2,rombach2022high}, such as stable diffusion, is essentially the same as the classifier-guidance, in which $\mu_{\theta}(x_t,t)+\sigma_t^2\bigtriangledown_{x_t}\log P(y|x_t)$ is included in $\mu_{\theta}(x_t,t,y)$.
\begin{equation}
P(x_{t-1}|x_{t},y) = \mathcal{N} (x_{t-1}, \mu_{\theta}(x_t,t,y),\sigma_{t}^2\mathcal{I}), 
\end{equation}

\begin{figure*}[!h]
  \includegraphics[width=\textwidth,height=0.55\textwidth]{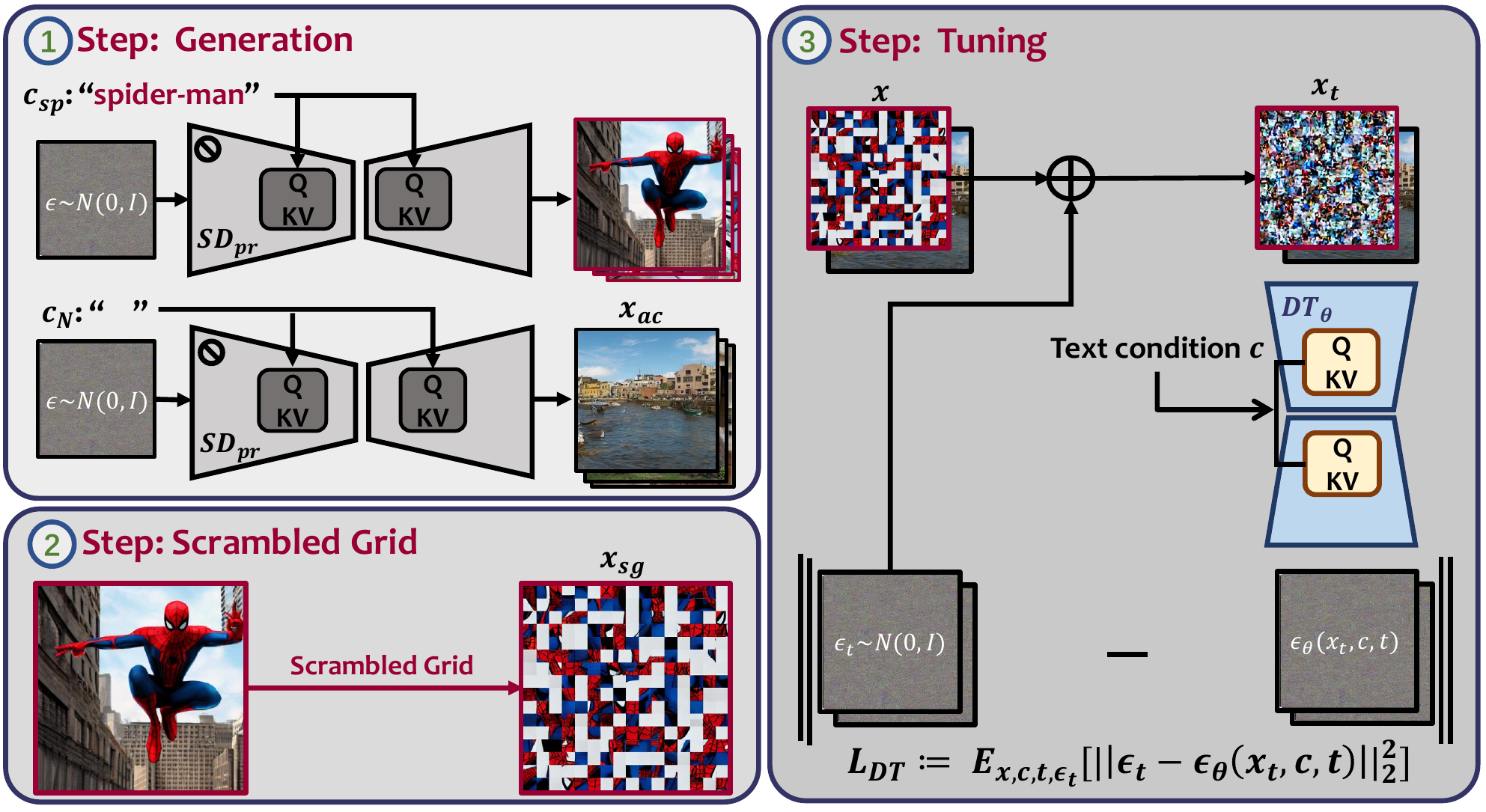}
  \vspace{-2em}
  \caption{The illustration of Degeneration-Tuning method can be divided into three steps. By using scrambled grid to reconstruct the correlation between unwanted concepts and its generative image domains, we guide the SD to generate meaningless contents when it receives conditional information about these textual concepts. }
  \label{main_process}
  \vspace{-0.5em}
\end{figure*}

\subsection{Degeneration-Tuning}
In Section \ref{Motivation}, we point out that $\Delta$ implicitly maps low-frequency information of the data domain. And stable diffusion has learned the low-frequency information of image content corresponding to textual concepts. As the low-frequency information of $x_0$ decreases, the distribution of $x_T$ converges to $\mathcal{N}\sim(0,\mathcal{I})$. This also indicates that the momentum of $\bigtriangledown_{x_t}\log P(y|x_t)$ approaches 0, and learning to fit the distribution of $x_0$ becomes easier for $\mu_{\theta}(x_t,t,y)$ than in the original case. Building upon this insight, we propose Degeneration-Tuning (DT) method. By fine-tuning the known conditional information of the pre-trained stable diffusion to point to a more readily fitted data distribution, the DT method effectively masks original semantic contents from specific textual concepts. 

The complete Degeneration-Tuning(DT) process has been shown in Figure \ref{main_process} and can be divided into three steps. The first step, "Generation," involves sampling Gaussian noise $\epsilon\sim\mathcal{N}(0,\mathcal{I})$ and using specific text conditions $c_{sp}$ to generate images that include the desired textual concepts (e.g., using "spider-man" as an example). 
Beside of conditional information $c_{sp}$, to avoid overfitting during tuning, we use a None condition $c_{N}$ (" ") to construct anchor images $x_{ac}$. The second step is "Scrambled Grid". Firstly, we grid images which generated by specific conditional information $c_{sp}$ and randomly reorder them to create scrambled images $x_{sg} = O(SD_{pr}(\epsilon,c_{sp}))$, where the $O$ refers to the Scrambled Grid operation and the size of grid in there is $16 x 16$. The reason for using this operation is that such a destruction technique can destroy low-frequency information in the original image while preserving some high-frequency features. 
The final step in degeneration-tuning process is "Tuning". Firstly, we construct tuning data $x \in {x_{sg} \bigcup x_{ac}}$ using $x_{sg}$ and $x_{ac}$. Next, we utilize this tuning data and their corresponding text conditions $c \in {c_{sp} \bigcup c_{N}}$ to fine-tune the parameters in $SD_{pr}$, which is represented by $DT_{\theta}$.
The training loss $L_{DT}$ becomes:
\begin{equation}
\begin{aligned}
L_{DT} :&= \mathbb{E}_{x, c, t, \epsilon _t \sim \mathcal{N}(0,1)}[ ||x-DT_{\theta}(\alpha_t x + \sigma_t\epsilon_t,c,t)||_2^2 ], \\
&= \mathbb{E}_{x, c, t, \epsilon _t \sim \mathcal{N}(0,1)}[ ||\epsilon_t - \epsilon_{\theta}(x_t,c,t)||_2^2]
\end{aligned}
\end{equation}

Since the Scrambled Grid operation narrows the distribution between $x_T$ and $\hat{x}_T$, training $DT_{\theta}$ to fit this specific data domain becomes easier comparatively. For a single concept in DT, it is sufficient to construct just 800 to 1000 $x_{sg}$ samples and and an equal number of $x_{ac}$ samples. The process of degeneration-tuning requires only a minimal learning rate of $1e^{-7}$, which is significantly lower than the learning rate used in traditional stable diffusion fine-tuning ($1e^{-4}$) as reported by \cite{rombach2021highresolution}, and demands a relatively low number of training epochs, approximately 60. The trained components in the SD is the entire U-net framework \cite{ronneberger2015u}.


\section{Experiments}
\label{experiments}
In this Section, we show the performance of our Degeneration-Tuning (DT) method in shielding various types of concepts. The pre-trained stable diffusion (SD) we utilized is SD-1.5 \cite{sd15}, which has been open-sourced in Huggingface \cite{hgface}. All of our training experiments were conducted on a single machine equipped with 8-GPU V100 GPUs. The batch size was set to 16, the learning rate was uniformly adjusted to $1e^{-7}$, and the training epochs were consistently set at 60. For each individual concept, we set the number of $X_{sg}$ and $X_{ac}$ samples at 900 and 1200, respectively. When masking multiple concepts simultaneously, it is sufficient to simply stack multiple degraded datasets together. 

\subsection{\textbf{Effectiveness in Recontextualization}}
\begin{figure}[!h]
  \includegraphics[width=0.47\textwidth,height=0.37\textwidth]{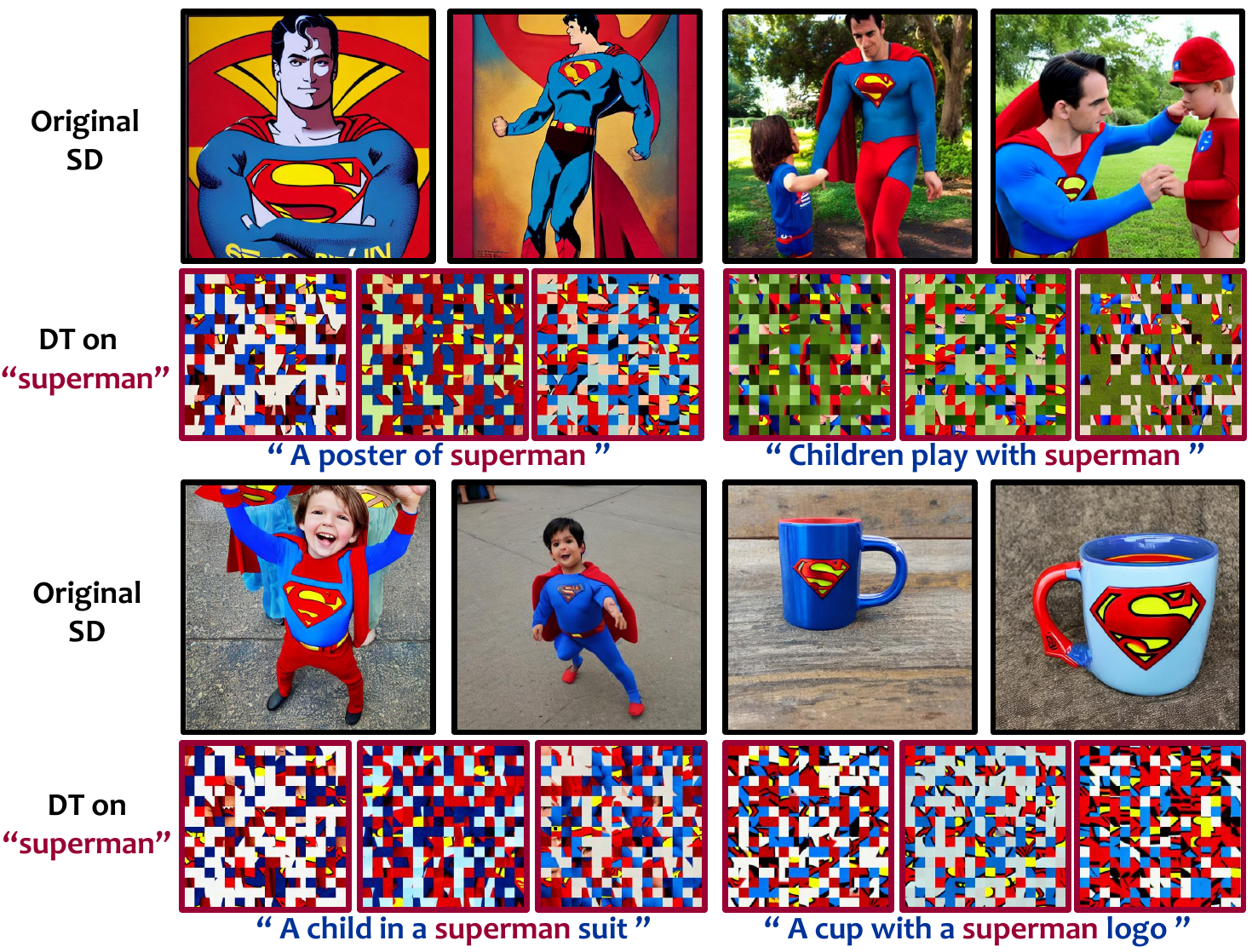}
  \vspace{-1em}
  \caption{The performance of the SD in different contexts containing the concept "superman" is demonstrated before and after applying DT, using 'superman' as the conditional information. 
  }
  \label{Recontextualization}
  \vspace{-1em}
\end{figure}

When we aim to shield a concept, our goal is not only to ensure the model can be activated when a single textual concept is inputted, but also to be effective in all contexts containing that concept. In DT, although we only target specific concepts as the model's shield goals, we find that it can still be effectively applied to various contexts containing these textual concepts. Taking the concept of "superman" as an example. After applying DT to the concept word "superman" (DT on "superman"), the model can shield content not only in single-object prompts such as "a photo of superman" but also in prompts containing multiple objects and meanings, like "children play with superman" and "A cup with a superman logo." The detailed results are presented in Figure \ref{Recontextualization}. By comparing the generated images of the original stable diffusion model and the model after DT on "superman", we demonstrate that our DT method is effective in recontextualization.

\subsection{\textbf{Effectiveness in protecting artistic styles}}
\begin{figure}[!h]
  \includegraphics[width=0.47\textwidth,height=0.37\textwidth]{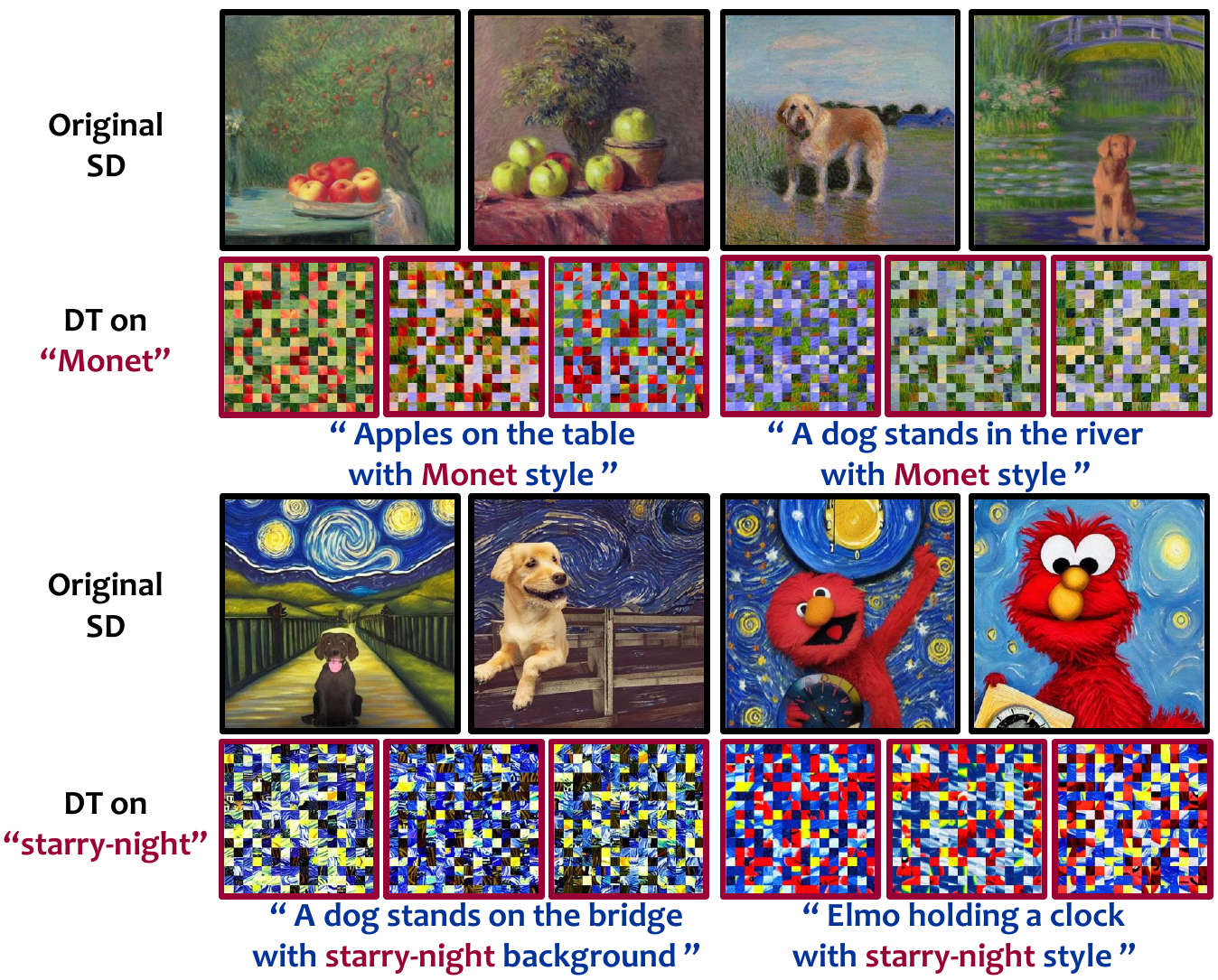}
  \vspace{-1em}
  \caption{The performance of the SD in generating styles characteristic of 'Monet' and 'starry-night', before and after DT on the textual concepts "Monet" and "starry-night". 
  }
  \label{artistic_styles}
\end{figure}

Imitation and plagiarism of artistic styles are the most common copyright problems in generative models. In fact, recent lawsuits against companies such as Stability AI, DeviantArt, and Midjourney highlight the seriousness of this issue \cite{qisu}. To demonstrate that our degeneration-tuning (DT) method remains effective in masking artistic styles, we applied it to the concepts of "Monet" and "Starry Night." The results, as displayed in Figure \ref{artistic_styles}, indicate that the DT method still excel in shielding the content of artistic styles.

\begin{figure}[!h]
  \includegraphics[width=0.47\textwidth,height=0.4\textwidth]{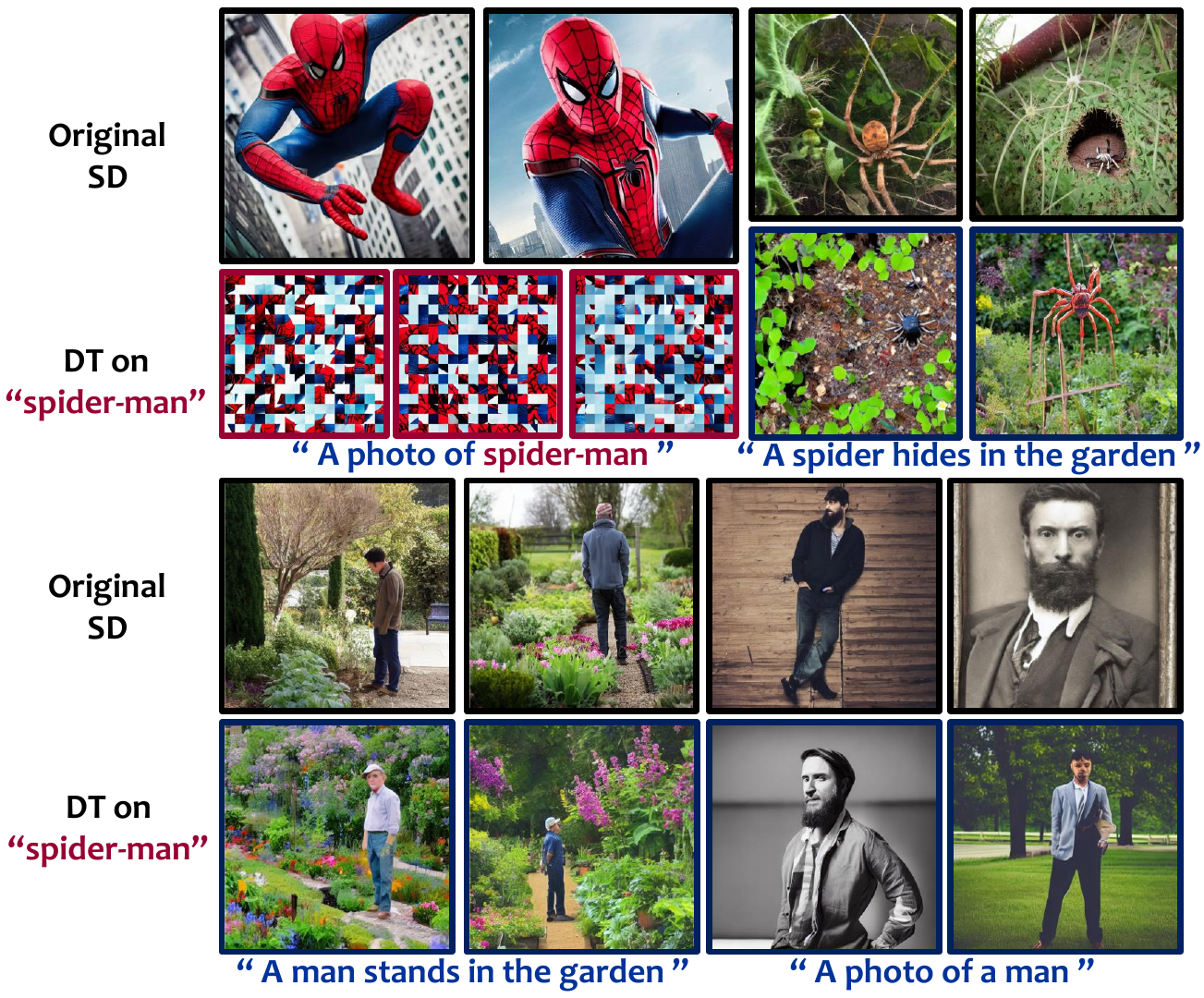}
  \vspace{-1em}
  \caption{The performance of the SD in concepts related to "spider-man" is shown both before and after DT on the textual concept "spider-man". 
  }
  \label{Tokenization}
  \vspace{-1em}
\end{figure}
\subsection{\textbf{Effectiveness in various other concepts}}
In addition to demonstrating the effectiveness of our DT method in shielding specific concepts, in there, taking the concept "spider-man" as an example, we showcase its generative performance on non-specific conceptual content. As shown in Figure \ref{Tokenization}, after DT on textual concept "spider-man" in SD, we can observe that the model does not exhibit significant deviation or degradation in generating content for the concepts of "spider" and "man". 
\subsection{\textbf{Effectiveness in Grafting}}
\begin{figure}[h]
  \includegraphics[width=0.47\textwidth,height=0.4\textwidth]{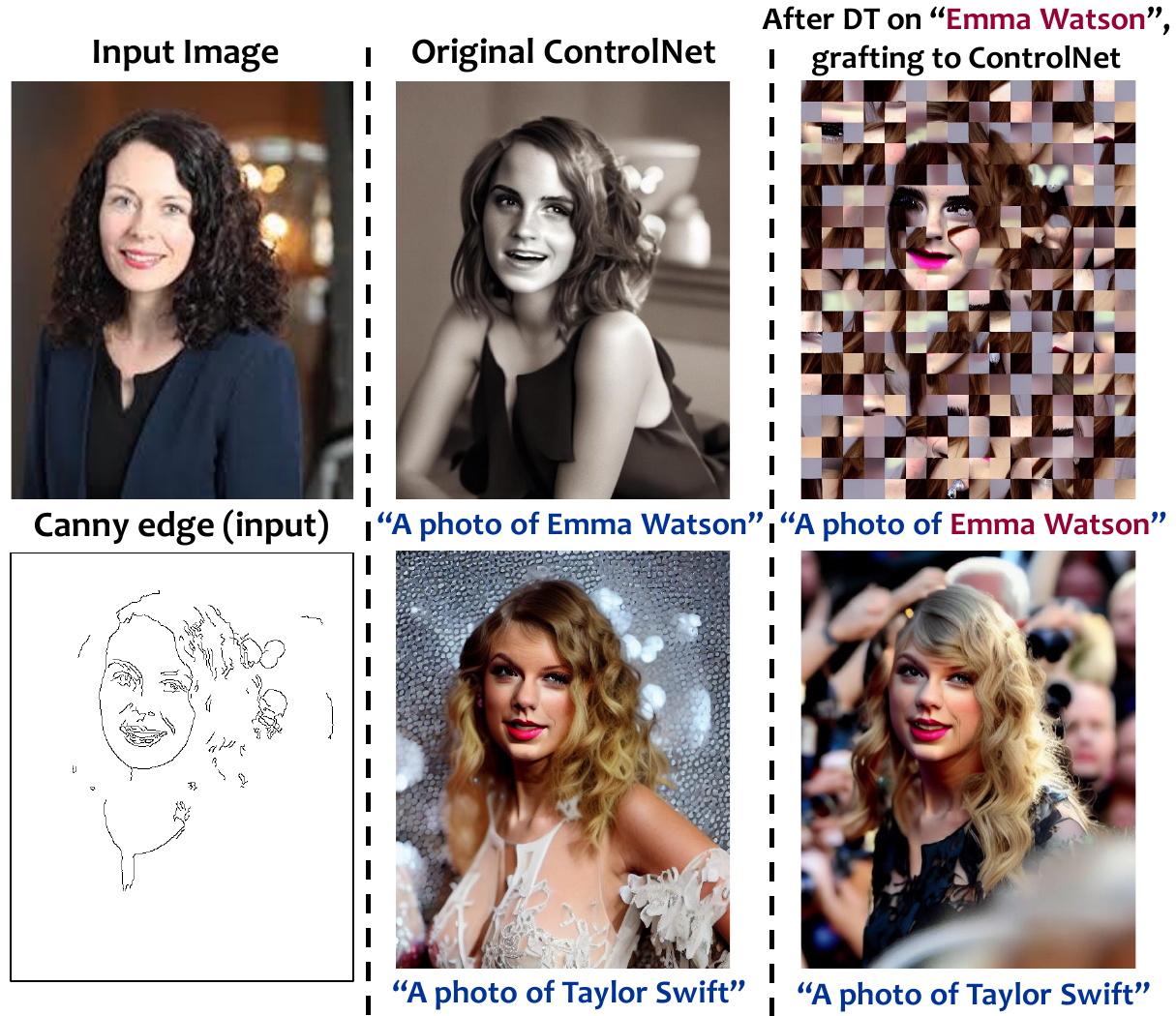}
  \vspace{-1em}
  \caption{The performance of the edge-based ControlNet in the concept of Emma Watson when given a canny edge input, before and after being grafted from the model that underwent DT on the text "Emma Watson".
  }
  \label{emmawatson}
  \vspace{-1em}
\end{figure}

In addition to the performance of the model after DT, we observe that the $DT_{\theta}$ model, which has been degeneration-tuned in specific textual concepts, demonstrates a grafting ability. Specifically, when we replace the U-net network of ControlNet with the model $DT_{\theta}$, this grafted ControlNet (Con-DT) is able to shield these specific concepts present in the model $DT_{\theta}$, even when additional conditional information such as pose and canny information is inputted along with the textual information. We demonstrate the model's grafting ability in Figures \ref{first_image} and \ref{emmawatson} for the concepts "spider-man" and "Emma Watson", respectively. In Figure \ref{first_image}, we show that the Con-DT can generate images based on both pose and text information, such as 'robot-man', while effectively shielding the content about 'spider-man', when we graft the U-net after DT on the 'spider-man' concept into a pose-based ControlNet. Similarly, in Figure \ref{emmawatson}, the performance of the edge-based Con-DT aligns with pose-based Con-DT. The edge-based Con-DT is able to generate content based on canny edge and text information, such as "a photo of Taylor Swift", while still shielding the content about 'Emma Watson'. We believe that the reason for this is that conditional information from different modalities refers to different semantic content, and the information spectra of these semantic content are not the same. Masking the textual concepts does not affect the influence of other modalities' conditional information on the generative content.

\subsection{Fine-tuning using normal images}
\begin{figure}[!h]
  \vspace{-1em}
  \includegraphics[width=0.47\textwidth,height=0.28\textwidth]{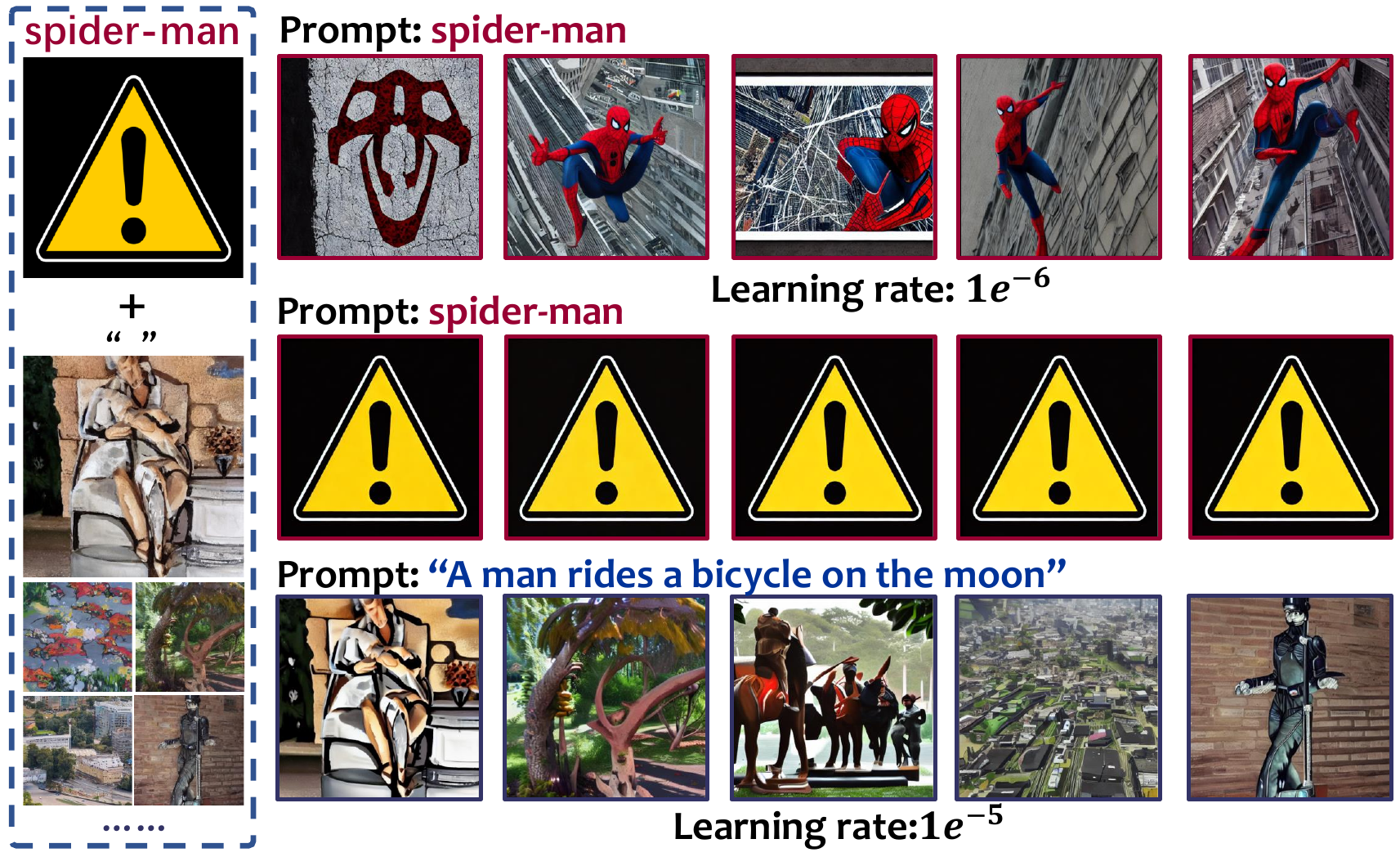}
  \vspace{-1em}
  \caption{The performance of DT on the text "spider-man" with different learning rates, without using the Scrambled Grid operation.}
  \label{without_SG}
  \vspace{-1em}
\end{figure}

In Section \ref{methods}, we noted that Scrambled Grid (SG) strategy is critical to the effectiveness of the degeneration-tuning method. In there, we show the results of the DT method without SG strategy. Taking the concept "spider-man" as an example, we replaced the data $x_{sg}$, which underwent SG, with specific content images, and combined them with images generated from none condition information (" ") to create the tuning dataset $x$. The results are shown in Figure \ref{without_SG}. Despite elevating the learning rate from 1e$^{-7}$ to 1e$^{-6}$ and extending the training epoch from 100 to 200, the fine-tuned model failed to alter its generative content when supplied with the condition text "spider-man" (first-row). When we increased the learning rate to 1e$^{-5}$, the model no longer generate images associated with the concept of "spider-man". However it also lost the ability to modulate the generated contents based on textual information (last-two-row). This situation indicates that the model has overfitted to the tuning dataset $x$, emphasizing the difficulties faced by the diffusion model in learning low-frequency signals as compared to high-frequency. 

\section{Evaluation}
\label{evaluation}

In this section, we adopt the evaluation metrics used in previous generative works \cite{rombach2022high,huang2022fastdiff,huang2022prodiff,huanggenerspeech}, using FID \cite{heusel2017gans} and IS \cite{szegedy2016rethinking} scores to assess the image generation quality of the model after degeneration-tuning (DT) in various types of concepts and multiple concepts simultaneously. The FID score measures the distance between the generated image distribution and the target distribution, with a smaller score indicating greater similarity. And IS score assesses image fidelity based on a pre-trained Inception-v3 model \cite{szegedy2016rethinking}, with a higher score indicating better quality. 
Beside of this, we discuss the feasibility and challenges of continual degeneration-tuning, which becomes relevant as the number of copyright-related content continues to increase.
\subsection{Quantitative analysis}
We present a quantitative evaluation of the image quality generated by the models after applying DT to different concepts, using FID and IS scores. To demonstrate the effectiveness of our degeneration-tuning method, we separately evaluate the model's ability to generate both unwanted conceptual images and other general images. For specific conceptual content, we establish the image distribution generated by the original stable diffusion using these textual concepts as the target distribution. Concurrently, we use the image distribution generated by the model after DT on these textual concepts as the generated distribution. For other general content, we align with previous works \cite{rombach2022high,kawar2022imagic,dalle2} utilizing prompts from the COCO-30k validation set \cite{lin2014microsoft} to generate images and comparing them to the original COCO-30k distribution. 

Firstly, we evaluate the changes in the image generation quality of the stable diffusion model before and after degeneration-tuning on a single concept. The detailed results can be seen in the Table \ref{FID_IS}, 
\begin{table}[!h]
\vspace{-1em}
\begin{center}
\begin{tabular}{c|c|c|c|c}
\toprule
\multirow{2}{*}{DT in} & \multicolumn{2}{c}{C.s.c} & \multicolumn{2}{c}{COCO 30K} \\
 & FID $\downarrow$ & IS $\uparrow$ & FID $\downarrow$  & IS $\uparrow$ \\
\midrule
 \textbf{Original SD} & $\setminus$ & $\setminus$ & \textbf{12.61} & \textbf{39.20} \\
\midrule
 "spider-man" & 385.38 & 1.77 & 12.64 & 38.77 \\
 "superman" & 371.64 & 1.63 & 12.53 & 38.94 \\
 "Monet" & 355.20 & 1.81 & 12.60 & 39.12 \\
 "starry-night" & 360.02 & 1.70 & 12.62 & 38.73 \\
 "Emma Watson" & 392.21 & 1.58 & 12.58 & 38.60 \\
 "Donald Trump" & 381.71 & 1.60 & 12.70 & 39.01 \\
 \midrule
 "Joint" & 391.54 & 1.73 & 13.04 & 38.25 \\
\bottomrule
\end{tabular}
\caption{The FID and IS scores of the model before and after DT for unwanted concepts in the concept-specific content and COCO 30K dataset.}
\label{FID_IS}
\end{center}
\vspace{-2em}
\end{table}
where the C.s.c is "Contents about specific concepts". 

When comparing the FID and IS scores for specific concepts, we can find that the content about this concepts, which has been degeneration-tuned in the pre-trained stable diffusion model, cannot be generated. Additionally, the results on COCO-30K suggest that our method has little impact on generated contents outside of the specific concepts. Furthermore, we perform DT on all of this concepts jointly, and the FID and IS scores for this joint DT application are presented in the last row of Table \ref{FID_IS}. By comparing the result of the model after DT with the original SD, we observe that the number of unwanted concepts in DT does not significantly affect the model's generation quality on other content. Specifically, the FID score for COCO-30K increased by a mere 0.23 points, and the IS score decreased by 0.62 points, reinforcing the effectiveness of our method without compromising the generation quality of other content.
\begin{table}[!h]
\label{compare_methods}
\vspace{-1em}
\begin{center}
\setlength\tabcolsep{7.9pt}
\begin{tabular}{c|c|c|c|c}
\hline
Method	& Venue	& FID $\downarrow$ $/$ IS $\uparrow$ & CLIP & R.p.l\\
\hline
SLD\cite{schramowski2023safe} & CVPR'23 & 18.76 $/$ 36.64 & 0.1594	& \XSolidBrush \\
\hline
Erase\cite{gandikota2023erasing} & Arxiv'23 & 17.27 $/$ 37.21 & 0.1586 & \CheckmarkBold \\
\hline
\textbf{DT} & \textbf{Our} & \textbf{13.04 $/$ 38.25} & \textbf{0.1572}	& \CheckmarkBold \\
\hline
Original & CVPR'22	& 12.61 $/$ 39.20 & 0.1561 & \XSolidBrush \\
\hline
\end{tabular}
\caption{Comparing the performance of the DT method with existing content protection methods, SLD \cite{schramowski2023safe} and Erase \cite{gandikota2023erasing}. 
}
\end{center}
\vspace{-2em}
\end{table}

In addition to evaluating the efficacy of the DT method, we also compared the performance and differences of the DT method with existing content protection methods, such as SLD \cite{schramowski2023safe} and Erase \cite{gandikota2023erasing}. As shown in the Table \ref{compare_methods}, the resulting average FID and IS scores, along with CLIP scores, clearly show that the DT method maintains a closer resemblance to the original content in terms of general content generation compared to other methods. Among these concurrent works, only DT and Erase methods effectively prevent the risks of uncontrolled generation in the event of model parameter leakage by modifying the model parameters.

\subsection{Continual degeneration-tuning}
\label{continual_DT}
If we consider the training process of stable diffusion from a probabilistic perspective, degeneration-tuning the parameters is tantamount to finding their most probable values given some data $D_{j}$, where the $D_{j} = D_{sg} \bigcup D$. The $D_{sg}$ is the data after scrambled grid operation, and $D$ is the data used to pre-train the SD. The probability equation can be rearranged to:
\begin{equation}
\log p(\theta|D_{j}) = \log p(D_{sg}|\theta) + \log p(\theta|D) - \log p(D_{sg}),
\label{continual_probalility_optimize}
\end{equation}
This problem is challenging to address in traditional continual tasks where the original data $D$ is unavailable. However, for generative models like SD, $D$ can be re-generated from the model itself. This is the reason why we need to generate $x_{ac}$ in DT (in Section \ref{methods}). 

In there, we evaluated the performance of continual degeneration-tuning in unwanted concepts. The detailed FID and IS scores has been shown in Table \ref{continual_DT}.
\begin{table}[!h]
\vspace{-1em}
\begin{center}
\begin{tabular}{c|c|c|c|c}
\toprule
\multirow{2}{*}{DT in} & \multicolumn{2}{c}{C.s.c} & \multicolumn{2}{c}{COCO 30K} \\
 & FID $\downarrow$& IS $\uparrow$ & FID $\downarrow$& IS $\uparrow$\\
\midrule
 \textbf{Original SD} & $\setminus$ & $\setminus$ & \textbf{12.61} & \textbf{39.20} \\
\midrule
 "spider-man"  $\downarrow$& 385.38 & 1.77 & 12.64 & 38.77 \\
 "superman"  $\downarrow$& 380.12 & 1.70 & 13.10 &  38.02 \\
 "Monet"  $\downarrow$& 371.62  & 1.75  &  13.63 &  37.14 \\
 "starry-night" $\downarrow$& 392.48  & 1.68  & 14.21  &  36.58 \\
 "Emma Watson"  $\downarrow$& 387.61 & 1.76  &  14.87 &  36.21 \\
 "Donald Trump"  $\downarrow$& 390.25  &  1.71 &  15.32 & 35.71  \\
 \midrule
 "Joint" & 391.54 & 1.73 & 13.04 & 38.25 \\
\bottomrule
\end{tabular}
\caption{The FID and IS scores of the model after continual DT in different concepts.}
\label{continual_DT}
\end{center}
\vspace{-2em}
\end{table}
where the C.s.c is "Contents about specific concepts". From the scores, we observe that although continual DT does not significantly affect the model's performance in shielding specific conceptual contents, its performance on other prompts is negatively impacted. Compared to the joint results, the final FID score of the model on COCO-30K increased from 13.04 to 15.32, while the IS scores decrease from 38.25 to 35.71. By examining the quality of the images generated by the model after each continual tuning phase, we find that continual training amplified the bias between the generated image quality and the original image, leading to a butterfly effect and resulting in a deterioration of the model's generated image quality. 

\section{Conclusion}
In this paper, we conducted experiments to demonstrate that the primary factor influencing the semantic contents generated by stable diffusion is the distance between the initial sampled Gaussian noise and the final diffusion distribution within the training data domain. Building on this insight, we proposed a new method \textbf{Degeneration Tuning(DT)} to shield unwanted concepts from the level of stable diffusion's weights. In addition to showing the performance of our DT method in shielding various types of concepts qualitatively, the quantitative comparison of SD before and after DT indicates that DT method did not significantly affect the generative quality of other contents. Finally, we evaluated the feasibility of continual DT and proposed potential reasons that may lead to a decrease in the quality of generated image by the contunal model.

\section*{Acknowledgements}
This work has been supported in part by the Zhejiang NSF (LR21F020004), the NSFC (No.62272411), Alibaba-Zhejiang University Joint Research Institute of Frontier Technologies, and Ant Group. 


\bibliographystyle{ACM-Reference-Format}
\balance
\bibliography{acmmm2023}

\newpage
\appendix
\section{Analysis and Ablation}
In this Appendix, we provide further analysis of the mechanism behind the Degeneration-Tuning (DT) method and experimental results. Furthermore, we showcase various ablation experiments based on the DT proposed in the Section Methods.
\subsection{The impact of the degeneration-tuning on the original model parameters}
\begin{figure}[!h]
  \vspace{-0.5em}
  \includegraphics[width=0.47\textwidth,height=0.25\textwidth]{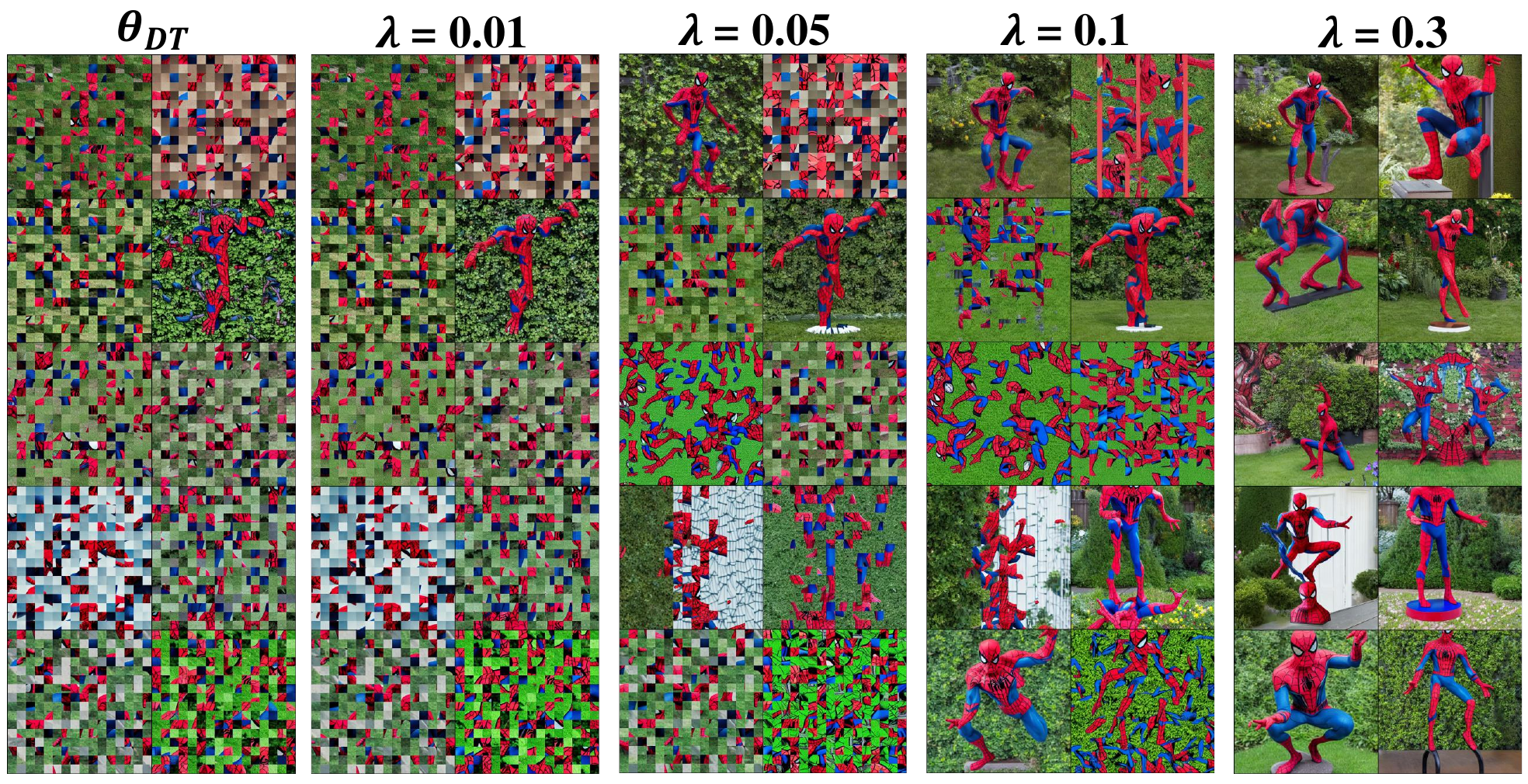}
  \vspace{-1em}
  \caption{The performance of fusion parameters generated by different hyper-parameter $\lambda$ on the prompt "spider-man stands in the garden".}
  \label{hyper_parameters}
  \vspace{-1em}
\end{figure}

In Section Methods, we mentioned that the DT method can fit the data processed by Scrambled Grid with a very small learning rate (1e-7). However, how much impact does it have on the original model parameters? Here we present an experiment to demonstrate it. Taking the concept "spider-man" as an example. Assuming the original pre-trained stable diffusion parameters are $\theta_{ori}$, and the model parameters after DT are $\theta_{DT}$. By performing a linear transformation, we obtain fused models $\theta_{f}$ under different hyper-parameter $\lambda$. where the $\theta_{f} = \lambda\theta_{ori}+(1-\lambda)\theta_{DT}$. Figure \ref{hyper_parameters} shows the masking effect of the model $\theta_{f}$ on the prompt "spider-man stand in the garden" under different hyper-parameter $\lambda$.
From the experiments, we can find that the shielding effect of the model on the textual concept "spider-man" starts to weaken when the hyper-parameter $\lambda$ changes from 0.01 to 0.05. When the hyper-parameter reaches 0.1, the model becomes sluggish towards the concept of spider-man. All of this demonstrates that although the influence of the DT on the original model parameters is small, its effect on the semantic contents of the generative image is significant.

\subsection{Applying DT to different modules of the model}
In Section Methods, we state that the DT method was used to adjust the parameters within the U-Net framework of stable diffusion. However, the U-Net framework includes the cross-attention modules and the resblock modules. Here, taking the concept "spider-man" as an example, we discuss the performance when applying DT only to the cross-attention modules or the resblock modules. 
\subsubsection{\textbf{Just applying DT to the cross-attention modules}}
\begin{figure}[!h]
  \vspace{-0.5em}
  \includegraphics[width=0.47\textwidth,height=0.23\textwidth]{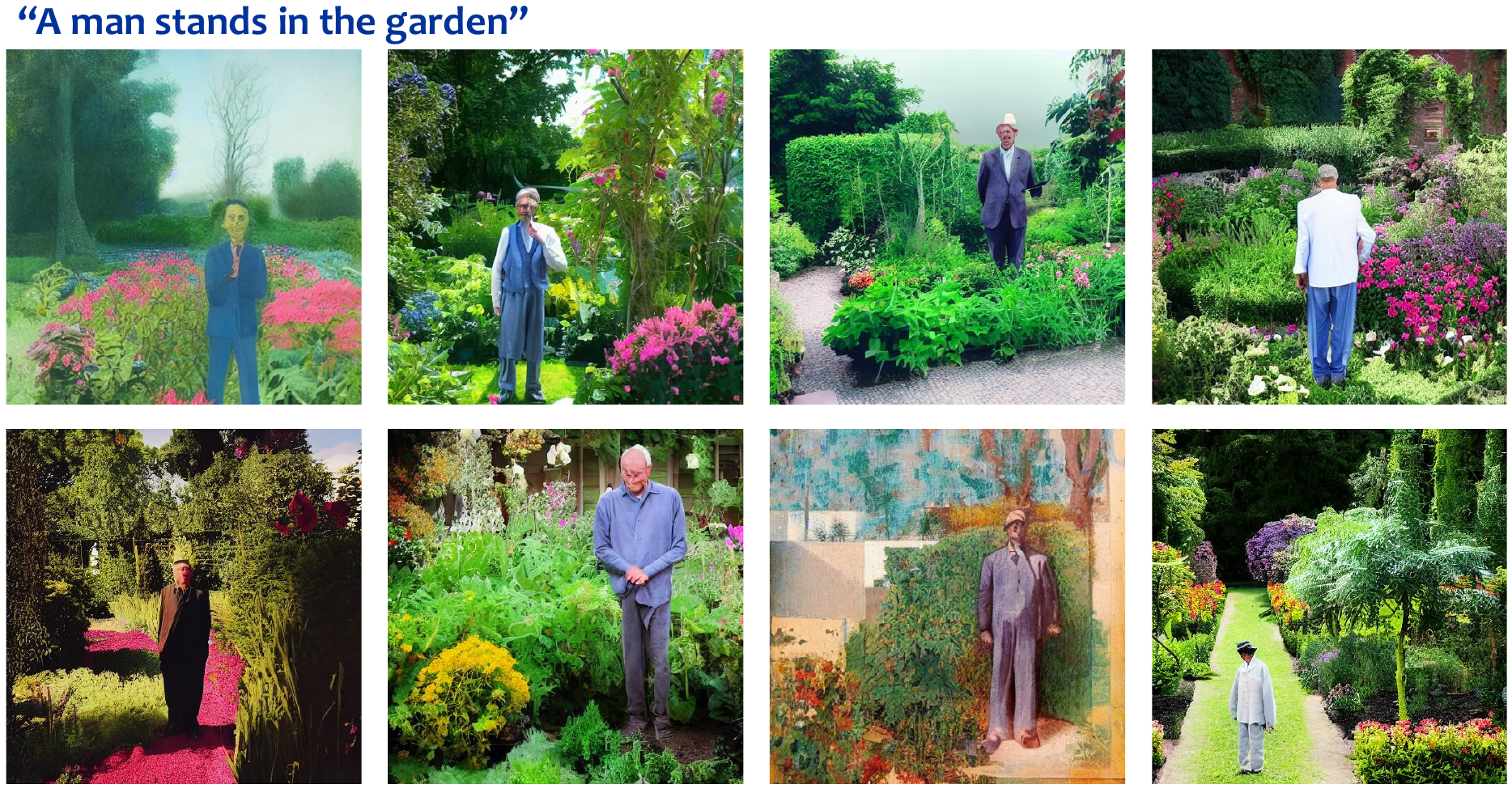}
  \vspace{-1em}
  \caption{The performance of the model, which applies DT only to the cross-attention modules, on the prompt 'A man stands in the garden'.}
  \label{just_attn}
  \vspace{-1em}
\end{figure}

After adjusting only the parameters of the cross-attention modules within the U-Net framework using DT with the textual concept of "spider-man", we show the performance of this model in the prompt "A man stands in the garden" without the concept "spider-man" in Figure \ref{just_attn}. We can clearly see that the content of these images contains textual information, but its saturation and contrast are very high, which makes the entire output appear unrealistic.

\subsubsection{\textbf{Just applying DT to resblock modules}}
\begin{figure}[!h]
  \vspace{-0.5em}
  \includegraphics[width=0.47\textwidth,height=0.23\textwidth]{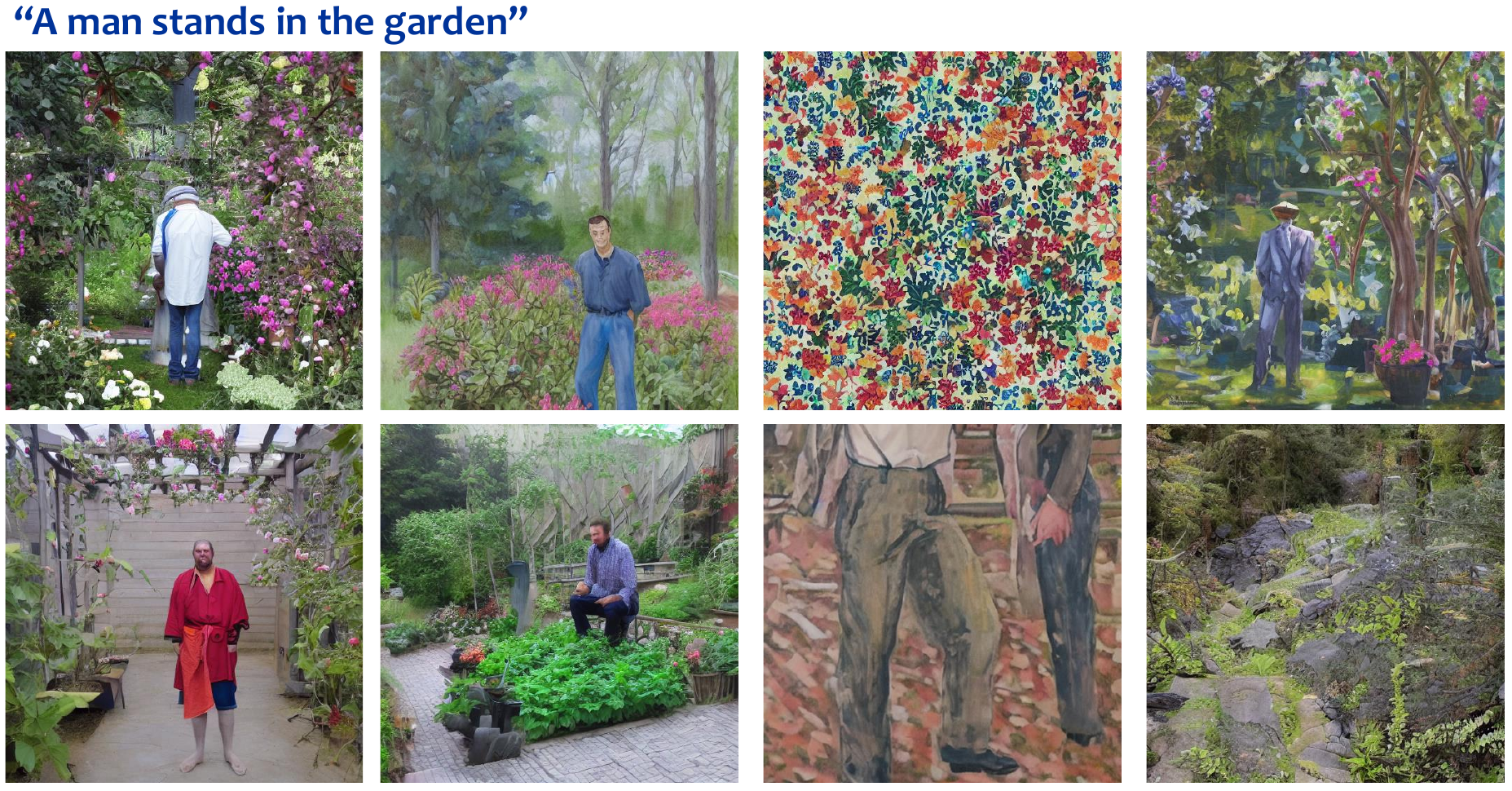}
  \vspace{-1em}
  \caption{The performance of the model, which applies DT only to the resblock modules, on the prompt 'A man stands in the garden'.}
  \label{no_attn}
  \vspace{-1em}
\end{figure}

After adjusting only the parameters of the resblock modules within the U-Net framework using DT with the textual concept of "spider-man", we show the performance of this model in the prompt "A man stands in the garden" without the concept "spider-man" in Figure \ref{no_attn}. We can clearly see that the content of these images is always different from textual information.

\begin{figure*}[!h]
  \vspace{-0.5em}
  \includegraphics[width=\textwidth,height=0.5\textwidth]{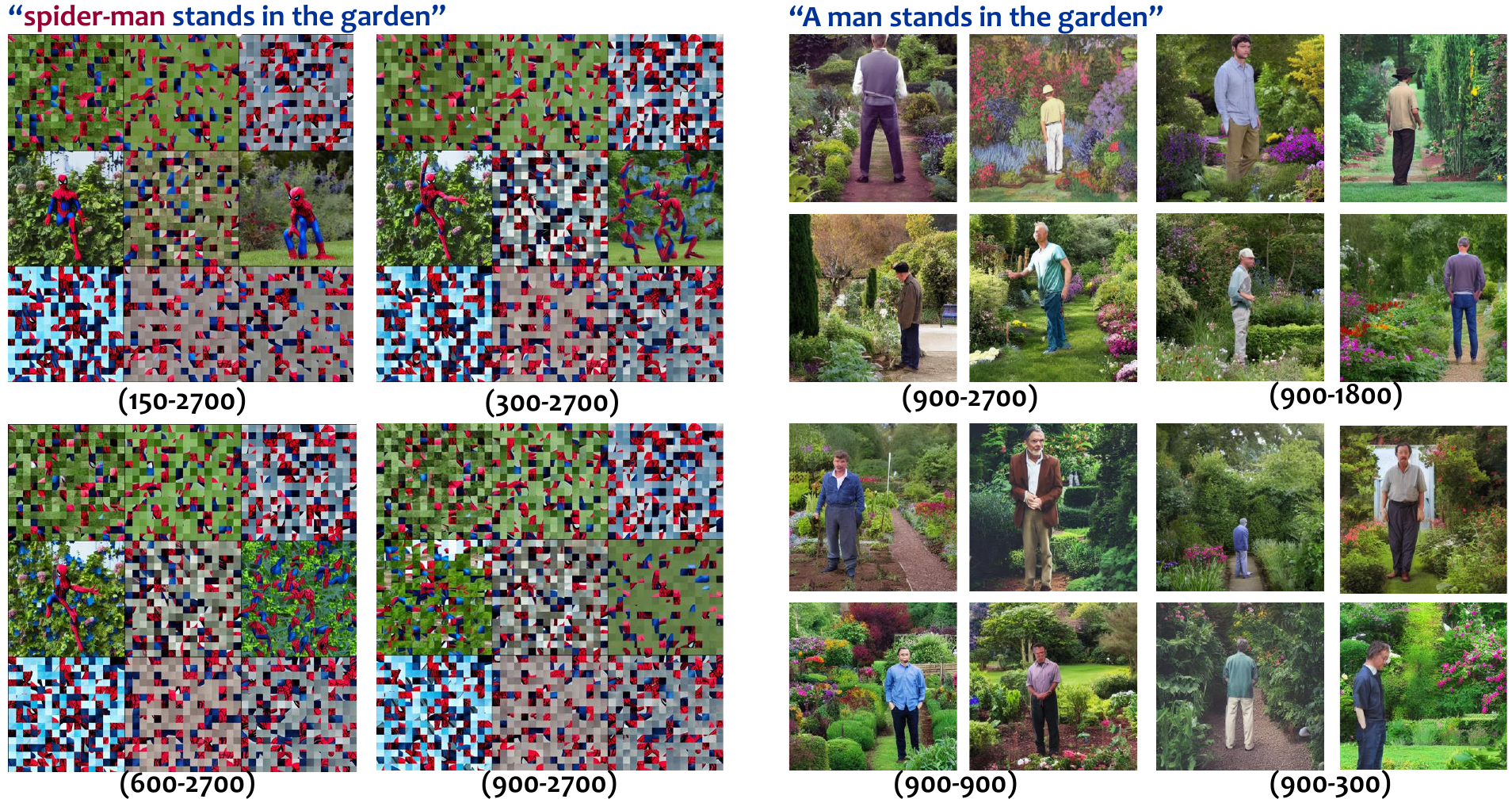}
  \vspace{-1em}
  \caption{Taking the concept "spider-man" as an example, we show the impact of different ratios and numbers of $x_{sg}$ and $x_{ac}$ images on DT performance.}
  \label{proportion}
  \vspace{-1em}
\end{figure*}
\subsection{Direct fine-tuning using pure color images}
\begin{figure}[!h]
  \vspace{-0.5em}
  \includegraphics[width=0.475\textwidth,height=0.23\textwidth]{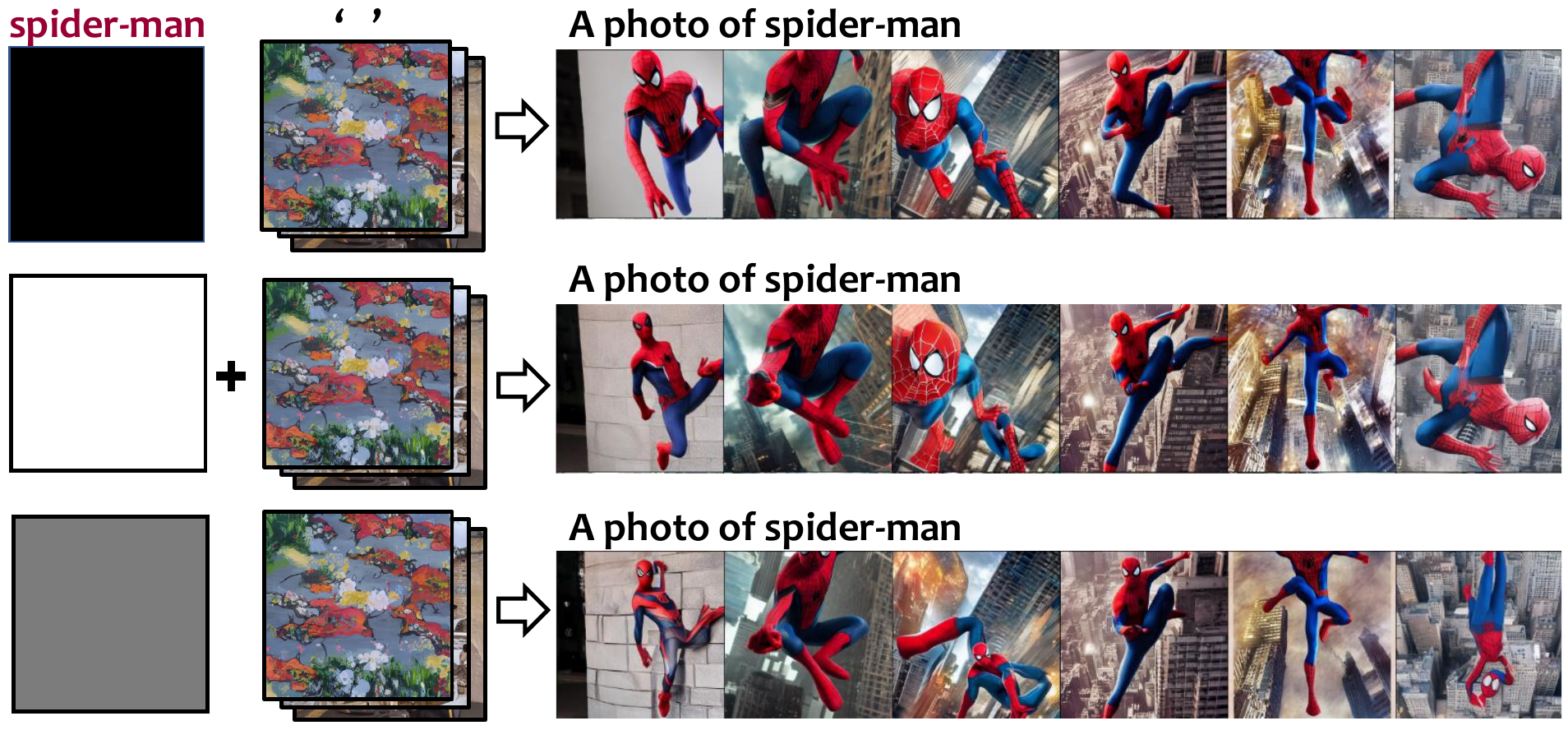}
  \vspace{-1em}
  \caption{The result of the model after fine-tuning on different replacement dataset.}
  \label{pure_color}
\end{figure}

In Section Experiments, we show the performance of DT without the Scrambled Grid operation. In this section, we supplement the results of using images with a single low-frequency signals (black, white and gray) instead of scrambled images in Figure \ref{pure_color}. From the results, we can see that regardless of which solid color image is used as a substitute for the text concept "spider-man", the model still generates images related to the concept "spider-man" when it receives conditional information about "spider-man". All of this once again indicates that the model's learning of low-frequency information is more difficult compared to high-frequency information. 

\subsection{The proportion of anchor images in Degeneration-Tuning}
In Section Experiments, we set the proportion of $x_{sg}$ to $x_{ac}$ to be approximately 1:1. In this section, taking the concept "spider-man" as an example, we show the impact of different ratios and numbers of $x_{sg}$ and $x_{ac}$ images on DT performance. From Figure \ref{proportion}, we can find that when the number of $x_{sg}$ increases from 150 to 900, its shielding effect on the concept "spider-man" varies. As the number of $x_{sg}$ increases, its effectiveness improves progressively. When we fix the number of $x_{sg}$ at 900 and adjust the quantity of $x_{ac}$, as shown in Figure \ref{proportion}, we find that for individual concepts, the performance of DT on other contents does not show significant changes. However, during the actual experimental process, we still set the number of $x_{ac}$ at 900$\sim$1200 to reduce the impact of DT on other contents. 

\subsection{The size of Scrambled Grid}
\begin{figure}[!h]
  \includegraphics[width=0.475\textwidth,height=0.13\textwidth]{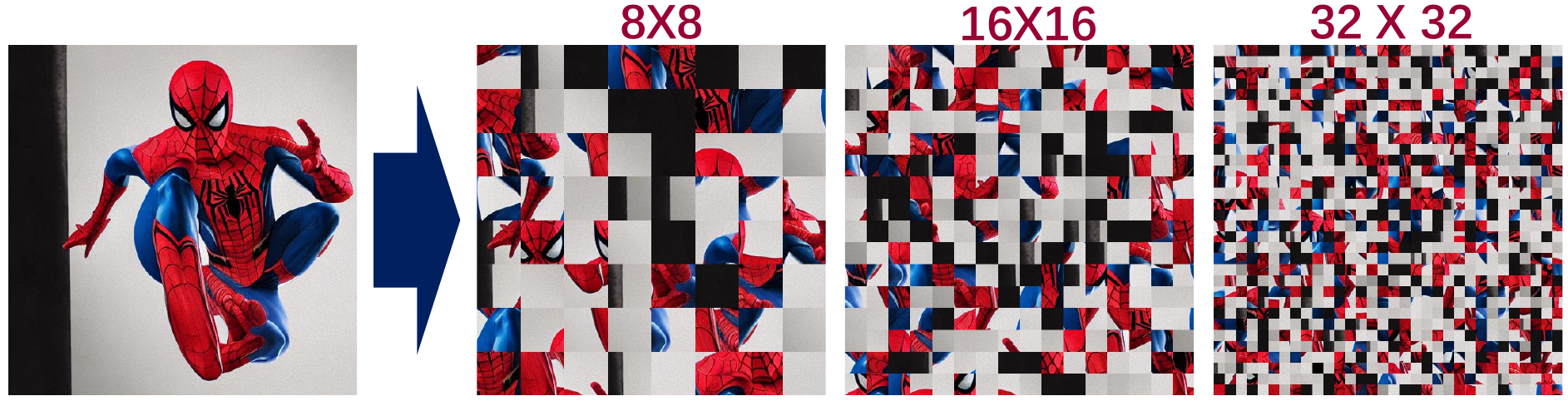}
  \vspace{-1em}
  \caption{Demonstration of Scrambled Grid with different grid size.}
  \label{size_SG}
  \vspace{-1em}
\end{figure}

In Section Methods, we stated that the size of Scrambled Grid in experiments were set to $16x16$. In this Section, we show the quantitative results of the model after DT with different size of Scrambled Grid on concept "spider-man". Figure \ref{size_SG} shows the demonstration of Scrambled Grid with different grid size. And Table \ref{FID_IS_size} shows the FID and IS scores of the model after DT with different size of Scrambled Grid on concept "spider-man". Based on the results, the influence of DT methods with different grid scales on the model's generation quality is not significant. However, the $8x8$ size visually retains too much image content, while the $32x32$ size destroys too much high-frequency information from the original content. Therefore, in practical experiments, we chose $16x16$.
\begin{table}[!h]
\begin{center}
\begin{tabular}{c|c|c|c|c}
\toprule
\multirow{2}{*}{The size of Scrambled Grid} & \multicolumn{2}{c}{C.s.c} & \multicolumn{2}{c}{COCO 30K} \\
 & FID & IS & FID & IS \\
\midrule
 \textbf{Original SD} & $\setminus$ & $\setminus$ & \textbf{12.61} & \textbf{39.20} \\
\midrule
 $8X8$ & 357.21 & 1.80 & 12.63 & 38.69 \\
 \textbf{$16X16$} & \textbf{385.38} & \textbf{1.77} & \textbf{12.64} & \textbf{38.77} \\
 $32X32$ & 384.62 & 1.72 & 12.59 & 38.72 \\
\bottomrule
\end{tabular}
\caption{The FID and IS scores of the model after DT with different size of Scrambled Grid on concept "spider-man".}
\label{FID_IS_size}
\end{center}
\vspace{-2em}
\end{table}

\begin{figure*}[b]
  \includegraphics[width=1\textwidth,height=0.48\textwidth]{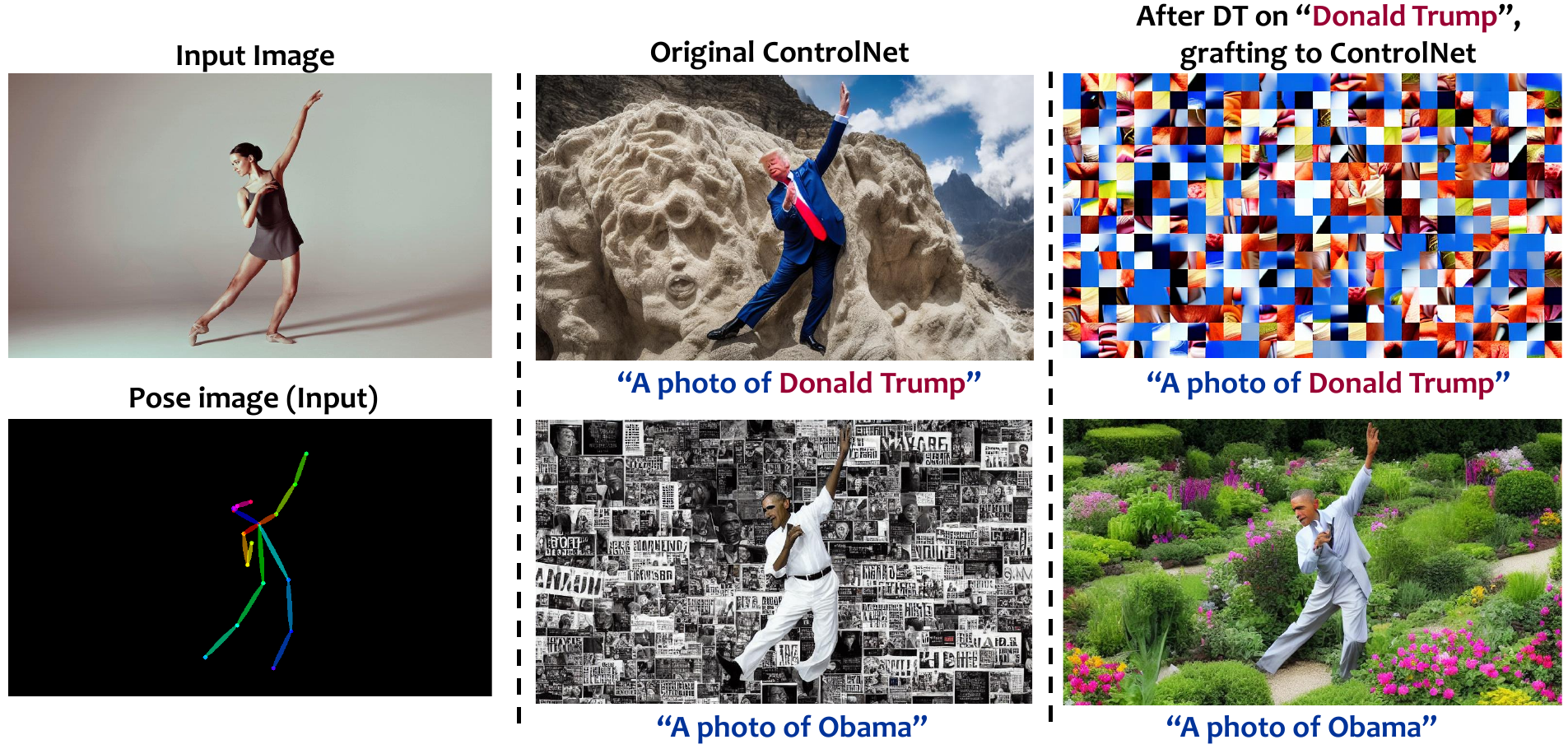}
  \vspace{-2em}
  \caption{The performance of the pose-based ControlNet in the content of Donald Trump when given a pose image input, before and after being grafted from the model that underwent DT on the textual concept "Donald Trump".}
  \label{donald_trump}
  \vspace{-0.5em}
\end{figure*}
\subsection{Another Performance of DT in Grafting}
We demonstrate the another performance model's grafting ability in Figures \ref{donald_trump} for the concepts "Donald Trump". When we graft the model after DT on concept "Donald Trump" into a pose-based ControlNet as shown in Figure \ref{donald_trump}, we show that Con-DT can generate images based on both pose and text information, such as "a photo of Obama", while still effectively shielding the content about "Donald Trump".

\subsection{The performance of the model on COCO 30K prompts after DT in multiple concepts.}
\begin{figure*}[!h]
  \vspace{-0.5em}
  \includegraphics[width=\textwidth,height=0.8\textwidth]{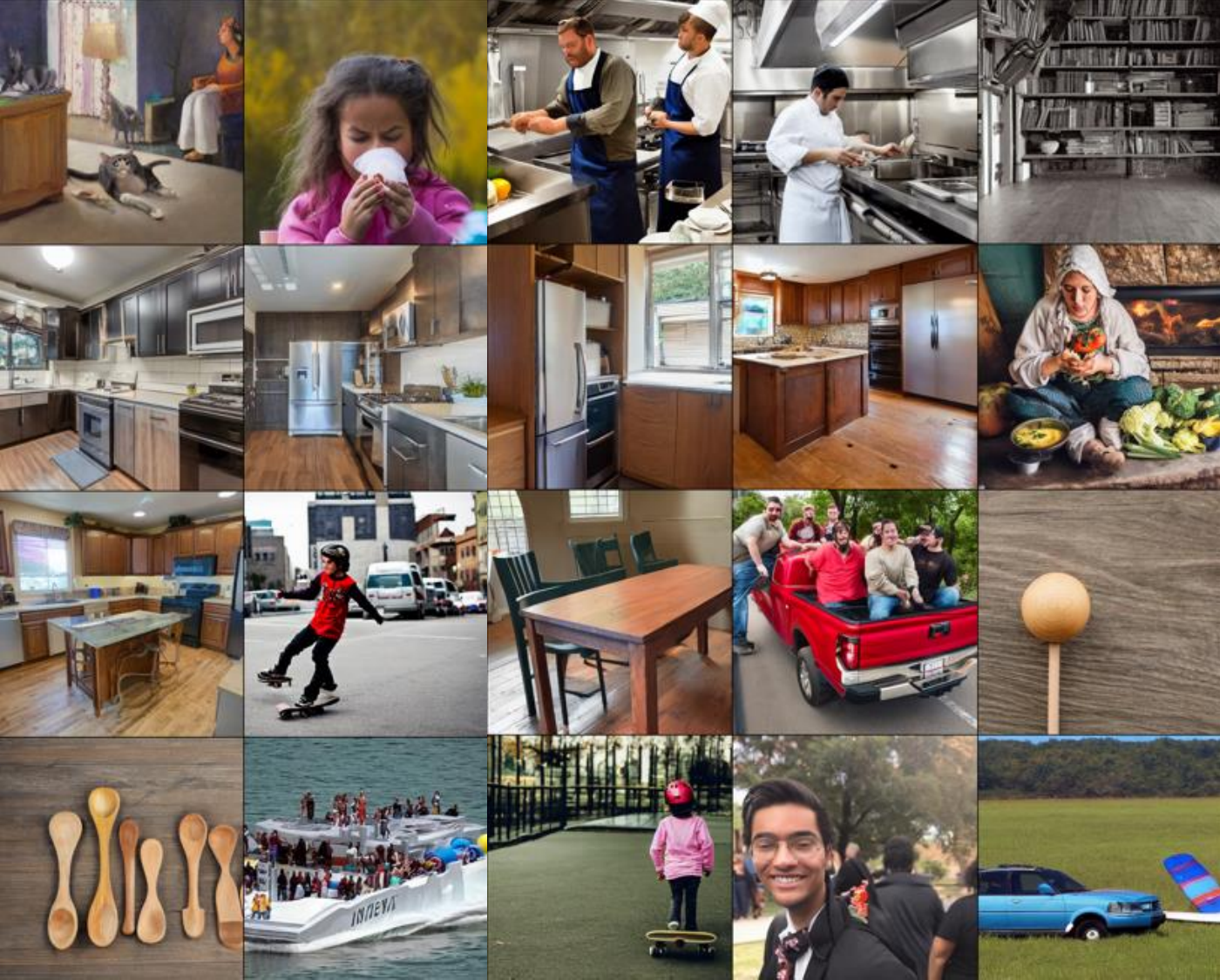}
  \vspace{-1em}
  \caption{The images generated by original stable diffusion on some COCO 30K prompts.}
  \label{sd15}
  \vspace{-1em}
\end{figure*}

\begin{figure*}[!h]
  \vspace{-0.5em}
  \includegraphics[width=\textwidth,height=0.8\textwidth]{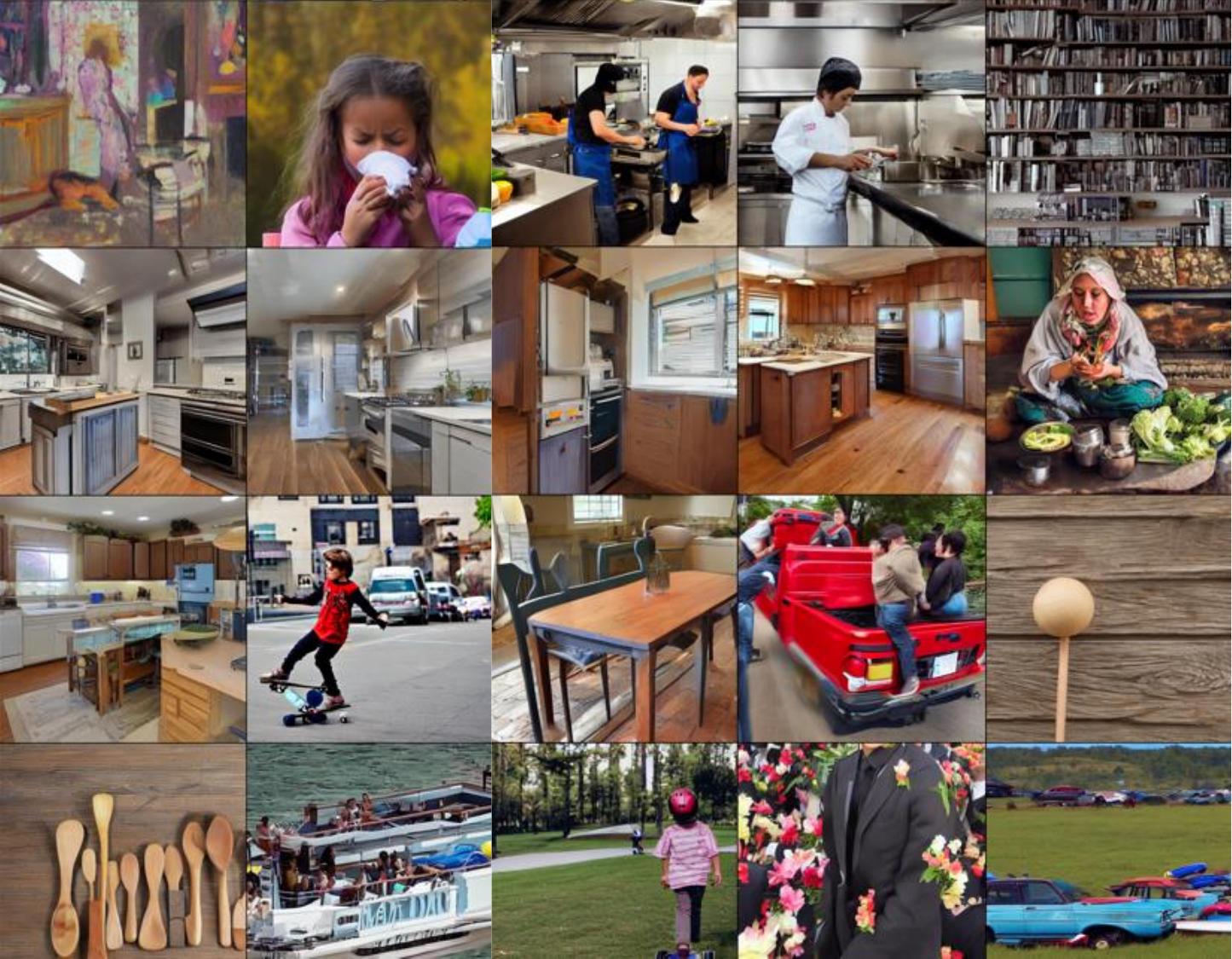}
  \vspace{-1em}
  \caption{The images generated by the model on some COCO 30K prompts after DT on multiple concepts.}
  \label{multi_ip}
  \vspace{-1em}
\end{figure*}

In Section Evaluation, we compared the quantitative results of the model before and after DT on COCO 30K prompts. In this Section, we show the generative images by the model before and after DT on COCO 30K prompts qualitatively. Figure \ref{sd15} shows the performance of the original stable diffusion on COCO 30K prompts. And Figure \ref{multi_ip} shows the performance of the stable diffusion on COCO 30K prompts after DT in multiple concepts.

\subsection{The performance of the model on COCO 30K prompts after continual DT on multi concepts.}
\begin{figure*}[!h]
  \vspace{-0.5em}
  \includegraphics[width=\textwidth,height=0.7\textwidth]{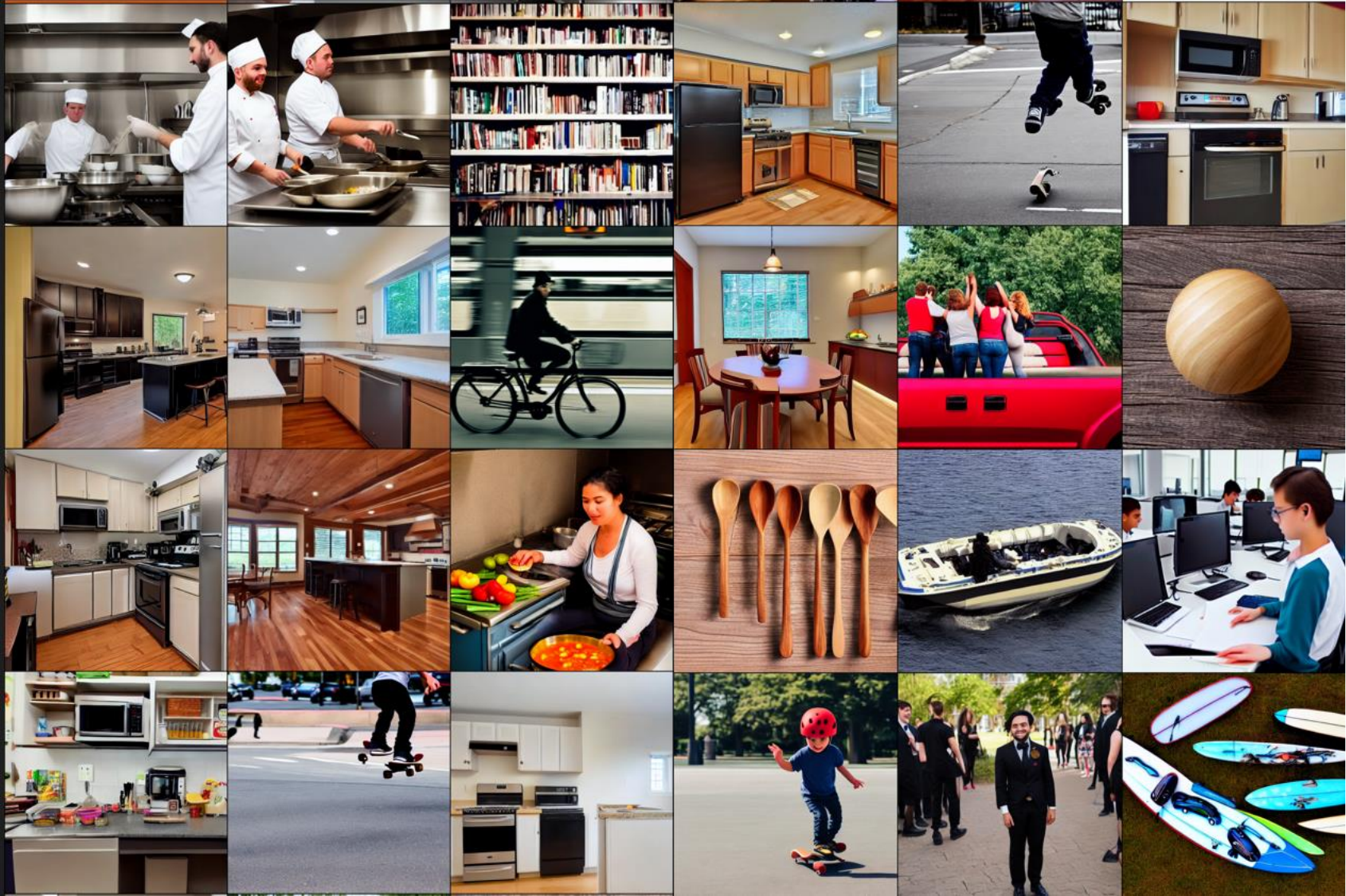}
  \vspace{-1em}
  \caption{The generative images by the model on some COCO 30K prompts after continual DT on multiple concepts.}
  \label{continual}
  \vspace{-1em}
\end{figure*}

\begin{figure*}[!h]
  \vspace{-0.5em}
  \includegraphics[width=\textwidth,height=0.8\textwidth]{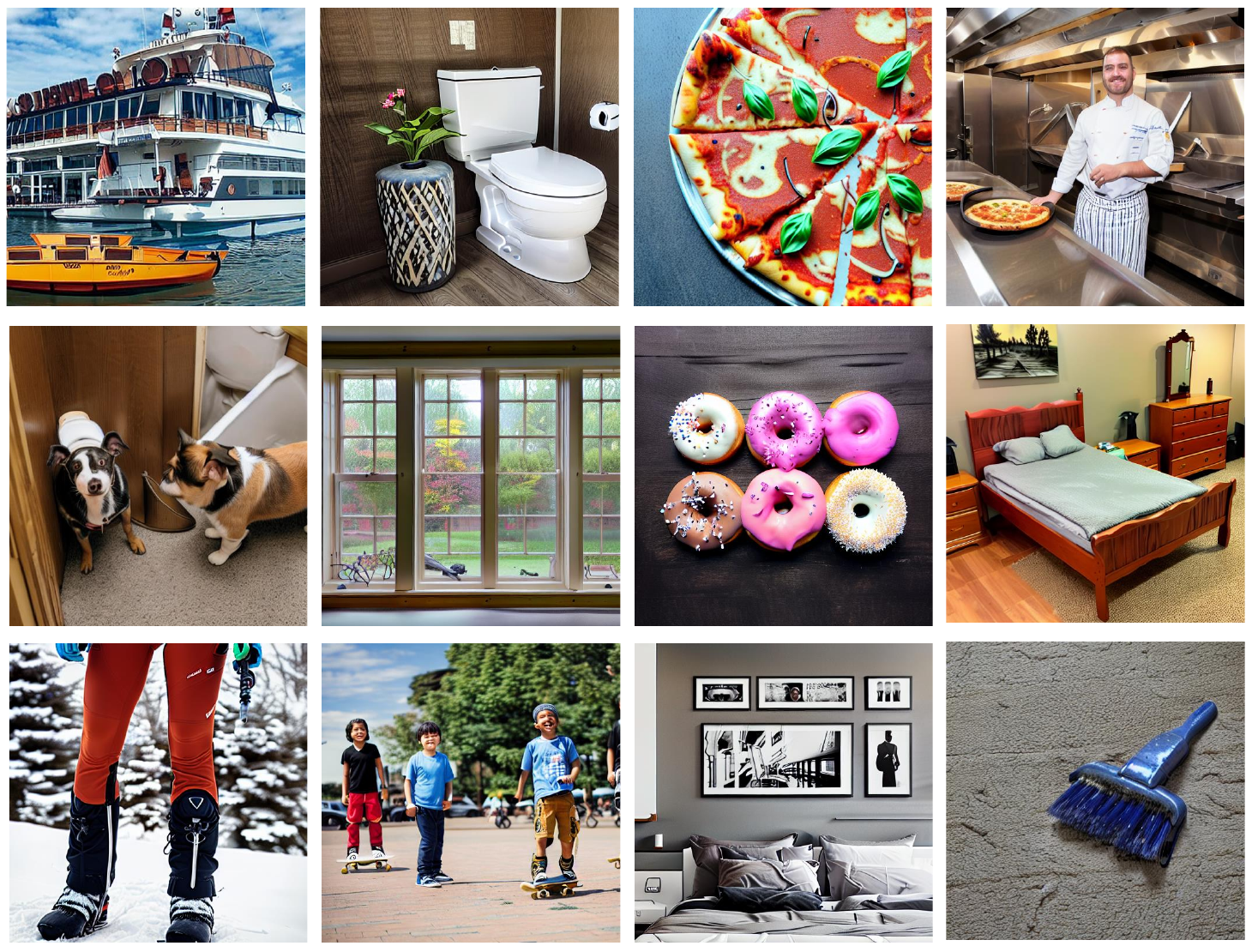}
  \vspace{-1em}
  \caption{The generative images by the model on some COCO 30K prompts after continual DT on multiple concepts.}
  \label{continual_2}
  \vspace{-1em}
\end{figure*}

In Section 5.2, we compared the quantitative results of the model before and after continual DT on COCO 30K prompts. In this Section, we show the generative images by the model on some COCO 30K prompts after continual DT on multiple concepts. Figure \ref{continual} and \ref{continual_2} show the performance of the this model on some COCO 30K prompts. 

\end{document}